\documentclass[10pt,twocolumn,letterpaper]{article}

\usepackage[pagenumbers]{cvpr} 

\definecolor{cvprblue}{rgb}{0.21,0.49,0.74}
\usepackage[pagebackref,breaklinks,colorlinks,allcolors=cvprblue]{hyperref}

\usepackage[normalem]{ulem}
\usepackage{multirow}
\usepackage{booktabs}
\usepackage{graphicx}
\usepackage{pifont}
\usepackage[dvipsnames]{xcolor}
\usepackage{colortbl}   
\usepackage[table]{xcolor} 
\newtheorem{observation}{Observation}
\newtheorem{definition}{Definition}
\newtheorem{remark}{Remark}
\newtheorem{theorem}{Theorem}
\newcommand{\Rom}[1]{\uppercase\expandafter{\romannumeral #1\relax}}
\usepackage[ruled]{algorithm2e}

\usepackage{amsmath}
\usepackage{amsfonts}
\usepackage{amssymb}

\def\paperID{376} 
\def\confName{CVPR}
\def\confYear{2026}

\title{ProxyFL: A Proxy-Guided Framework for Federated Semi-Supervised Learning}
\author{Duowen Chen \quad Yan Wang\footnotemark[1]\\
\normalsize{Shanghai Key Laboratory of Multidimensional Information Processing, East China Normal University}\\
{\tt\small duowen\underline{ }chen@hotmail.com, ywang@cee.ecnu.edu.cn}\\
\small{\url{https://github.com/DuowenC/FSSLlib}}
}

\begin{document}
\maketitle

\renewcommand*{\thefootnote}{\fnsymbol{footnote}}
\setcounter{footnote}{1}
\footnotetext{Corresponding Author.}
\renewcommand*{\thefootnote}{\arabic{footnote}}
\renewcommand*{\thefootnote}{\fnsymbol{footnote}}

\begin{abstract}
Federated Semi-Supervised Learning (FSSL) aims to collaboratively train a global model across clients by leveraging partially-annotated local data in a privacy-preserving manner. In FSSL, data heterogeneity is a challenging issue, which exists both across clients and within clients. External heterogeneity refers to the data distribution discrepancy across different clients, while internal heterogeneity represents the mismatch between labeled and unlabeled data within clients.
Most FSSL methods typically design fixed or dynamic parameter aggregation strategies to collect client knowledge on the server (external) and / or filter out low-confidence unlabeled samples to reduce mistakes in local client (internal). But, the former is hard to precisely fit the ideal global distribution via direct weights, and the latter results in fewer data participation into FL training. To this end, we propose a proxy-guided framework called \textup{ProxyFL} that focuses on simultaneously mitigating external and internal heterogeneity via a unified proxy. I.e., we consider the learnable weights of classifier as proxy to simulate the category distribution both locally and globally. For external, we explicitly optimize global proxy against outliers instead of direct weights; for internal, we re-include the discarded samples into training by a positive-negative proxy pool to mitigate the impact of potentially-incorrect pseudo-labels. Insight experiments \& theoretical analysis show our significant performance and convergence in FSSL.
\end{abstract}

\begin{figure}
    \centering
    \includegraphics[width=1\linewidth]{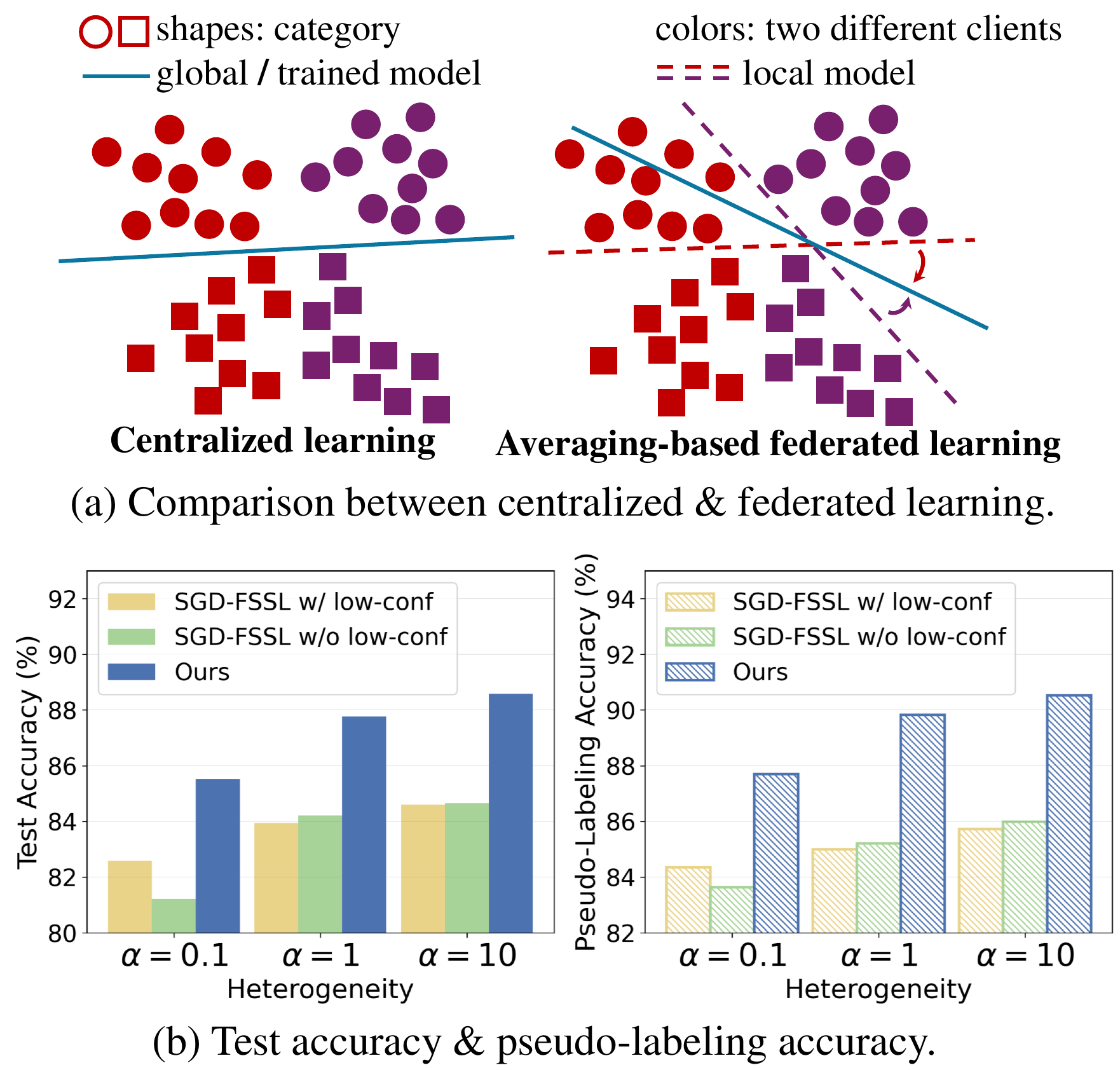}
    \caption{
    (a) Illustration of centralized learning and averaging-based federated learning. (b) Differences of test accuracy and pseudo-labeling accuracy under varying levels of heterogeneity (smaller $\alpha$ indicates greater heterogeneity). During each communication round, all clients are trained based on FedSGD \cite{mcmahan2017fedavg} \emph{w/}, \emph{w/o} low-confidence samples and our method for one local epoch.
    }
    \label{fig:intro}
\end{figure}

\section{Introduction}
\label{sec:intro}
The rapid advancement of edge devices and the Internet of Things (IoT) has led to a pressing need for decentralized training paradigms \cite{hoofnagle2019eureg,lim2020fededgesurvey}. Federated learning (FL), a distributed machine-learning paradigm, facilitates multi-device collaborative learning without sacrificing data privacy, which shares only model updates rather than raw data \cite{mcmahan2017fedavg}. Most existing FL works assume that local data in clients are fully labeled, but this assumption does not hold in practical scenarios when data annotation is laborious, time-consuming, or expensive. To remedy these issues, Federated Semi-Supervised Learning (FSSL) has emerged, enabling clients to train models leveraging both limited labeled data and a large amount of unlabeled data, thereby improving the performance of global model. In FSSL, data heterogeneity exists both across clients (external heterogeneity) and within clients (internal heterogeneity). The former refers to the distribution discrepancy across different clients, while the latter arises from the local mismatch due to 1) imbalanced sample sizes across different categories; 2) distribution imbalance between labeled \& unlabeled data.

For \textbf{external heterogeneity}, most studies design the dynamic aggregation weights to mitigate the imbalance across clients, based on relative sizes of local datasets \cite{bai2024feddure,liu2025sage} or some implicit statistics of local samples \cite{cho2023fedlabel,zhu2024feddb}, which may deviate from the ideal global distribution; For \textbf{internal heterogeneity}, some methods \cite{cho2023fedlabel,zhu2024feddb,liu2025sage} typically filter high-confidence unlabeled samples for training and exclude low-confidence ones to avoid pseudo-label bias, or dynamically assign a smaller weight to low-confidence samples, \emph{e.g.}, FedDure \cite{bai2024feddure}. However, these methods either lead to fewer data participation due to discarding low-confidence data or compromise on incorrect pseudo-labels only via a smaller weight. These insights motivate us to pose the following questions: \ding{172} \emph{Is there a better way than the aggregation weights to fit the global distribution without privacy concerns?} \ding{173} \emph{How can the local model leverage low-confidence unlabeled samples more effectively?}

To this end, we conduct some investigations into the above questions. As illustrated in Fig.~\ref{fig:intro}(a), the averaging-based aggregation strategy may lead to model deviation from the global category space, due to the distribution heterogeneity across clients; Fig.~\ref{fig:intro}(b) show test accuracy \& pseudo-labeling accuracy under {different} levels of heterogeneity. We observe that, as heterogeneity varies, training \emph{w/} and \emph{w/o} low confidence samples exhibit opposite trends to model performance. \emph{I.e.}, with {greater} heterogeneity (\emph{e.g.}, $\alpha=0.1$), the model ability is very limited, so \emph{w/ low-conf} outperforms \emph{w/o low-conf} due to the inclusion of more training samples. But, when the labeling ability of the model is more reliable {under} lower heterogeneity (\emph{e.g.}, $\alpha=1$ or $10$), incorporating extra low-confidence samples (\emph{w/ low-conf}) may hinder the model's category learning due to the errors from low-confidence pseudo-labels. {In short}, the performance of the two methods (\emph{w/} and \emph{w/o low-conf}) is inconsistent across different levels of heterogeneity. Neither method shows clear superiority over the other. So, it is crucial to consider how to utilize low-confidence samples.

To tackle Question \ding{172} and \ding{173} of the above observations in FSSL, we propose a new method called \textbf{ProxyFL} (\textbf{Proxy}-Guided \textbf{F}ederated Semi-Supervised \textbf{L}earning) that \textbf{leverages a unified proxy to {simultaneously} mitigate both internal and external heterogeneity}. \emph{I.e.}, we consider the learnable weights of classifier as proxies to model the category distribution both locally and globally. \textbf{Proxy does not sacrifice data privacy and bring negligible extra communication costs since the proxy itself is part of the model parameters in FL.} Firstly, we introduce a \textbf{G}lobal \textbf{P}roxy-\textbf{T}uning (GPT) mechanism, which explicitly optimize global proxies to fit the category distribution across clients, mitigating distribution shift from external heterogeneity. Secondly, we incorporate low-confidence unlabeled samples via a dynamic \textbf{I}ndecisive-\textbf{C}ategories \textbf{P}roxy \textbf{L}earning (ICPL) mechanism. For each low-confidence sample, we propose an \emph{indecisive-categories set} to represent its several possible categories instead of a single pseudo-label; For high-confidence unlabeled samples or labeled samples, we utilize the \emph{pseudo-label} or \emph{ground-truth}, respectively. Then we propose a relationship pool between unlabeled and labeled samples, and effectively train all samples based on the pool to mitigate internal heterogeneity. Experiments show that ProxyFL can significantly boost the performance and convergence of the FSSL model. 

The main contributions of this paper are as follows: 
\begin{itemize}
    \item To our best knowledge, this paper is the first to propose a unified proxy to mitigate both internal and external heterogeneity in FSSL. Note that our proxy does not bring privacy concerns or extra communication costs. 
    \item This paper proposes an FSSL method, ProxyFL, that can not only reduce the bias of averaging-based aggregation weights via an explicit optimization objective, but also effectively build the category relationship among all samples to facilitate more data participation. 
    \item This paper outperforms existing FSSL methods across multiple datasets. Extensive empirical \& theoretical analysis also show the effectiveness and convergence of our method under different levels of heterogeneity.
\end{itemize}

\section{Related Work}
\paragraph{Federated Learning}
Federated learning (FL) is a distributed machine learning approach that focus on safeguarding data privacy. One of the most challenging issues in FL is data heterogeneity, \emph{i.e.}, the distributions across clients are \underline{non}-\underline{i}ndependent and \underline{i}dentically \underline{d}istributed (non-i.i.d). To this end, various FL methods have been proposed to address this challenge such as regularization \cite{li2020fedprox}, additional data sharing \cite{tan2022fedproto}, aggregation strategies \cite{shi2025fedawa}, and personalization \cite{li2024fedotp}. For instance, FedProx \cite{li2020fedprox} requires each client to regularize with the previous global parameters to reduce local bias. However, these fully-supervised FL approaches struggle to generalize well under the scenarios of annotation scarcity. To address this issue, Federated Semi-Supervised Learning (FSSL) emerges, which enables clients to train local models via limited labeled data \& unlabeled data.
\paragraph{Semi-Supervised Learning}
Semi-Supervised Learning (SSL) aims to effectively leverage both limited labeled data and a large amount of unlabeled data to improve model performance. Two commonly-used strategies in SSL are consistency regularization and pseudo-labeling, respectively. Consistency regularization is based on the assumption that a model's prediction should remain consistent despite diverse perturbations to the inputs or model \cite{chen2021cps,yun2019cutmix,olsson2021classmix}. Another common strategy is pseudo-labeling, which determines pseudo-labels for unlabeled samples based on the high-confidence predictions of from the pre-trained model by labeled data and filter out low-confidence unlabeled samples \cite{sohn2020fixmatch,zhang2021flexmatch,wang2022freematch}. However, the pseudo-labels of these methods heavily depend on the confidence scores of model predictions, and if simply transferring them to the FL field, the number of local data will further decline due to the exclusion of low-confidence unlabeled samples.
\paragraph{Federated Semi-Supervised Learning}
Federated Semi-Supervised Learning (FSSL) addresses the challenge of training models on decentralized data where annotations are scarce. The field is often categorized into three scenarios: 1) Labels-at-Server, where clients only have unlabeled data and a central server holds labeled data \cite{diao2022semifl,he2021ssfl,jeong2020fedmatch,kim2023navigating,yang2024exploressl}. 2) Label-at-All-Client, where every client has a small fraction of labeled data and a large amount of unlabeled data \cite{jeong2020fedmatch,zhao2022fedgan}. 3) Labels-at-Partial-Clients, where a few clients are fully labeled while others are unlabeled \cite{li2023cbafed,liang2022rscfed,liu2021fdmedmatch,liu2024fedcd,zhang2024robust}. Our work focuses on the Label-at-All-Client setting, where a small fraction of samples have the ground-truths labels in each client, following  \cite{bai2024feddure,liu2025sage}. 

\section{Problem Formulation}
\label{sec:problem_statement}
This paper focuses on data heterogeneity in FSSL, including not only \textbf{external heterogeneity}\footnote{Definitions of FSSL data heterogeneity in Appendix~\ref{appendix:def_of_dh}. \label{foot:def}} across clients but also \textbf{internal heterogeneity}\textsuperscript{\ref{foot:def}} between labeled and unlabeled distributions within each client. Specifically, we assume that a federation system $\mathbb{C}$ consists of $K$ clients, denoted as $\mathbb{C} =\{\mathcal{C}_1, \ldots, \mathcal{C}_K\}$. For each $\mathcal{C}_k$, its local model is parameterized by $\mathbf{\Theta}_k$, which comprises a feature extractor $f_k$ parameterized by $\theta_k$, projecting local data to an embedding space $\mathbb{R}^d$, and a classifier $h_k$ parameterized by $\omega_k$, mapping the embedding space to category space $\mathbb{R}^C$, where $C$ indicates total category number. \emph{I.e.}, $\mathbf{\Theta}_k = \theta_k \cup \omega_k$. 

\begin{figure}
    \centering
    \includegraphics[width=0.9\linewidth]{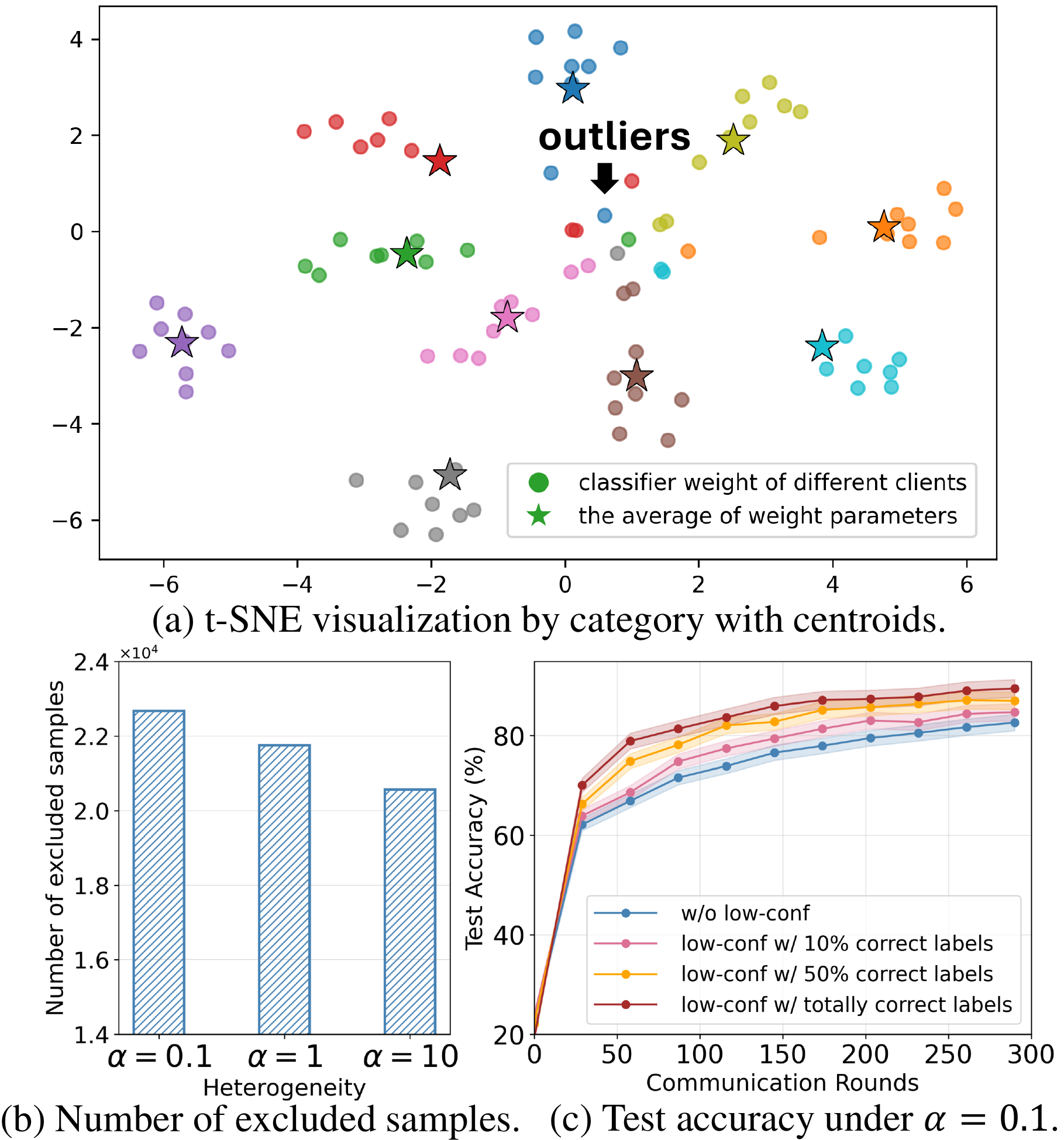}
    \caption{
    (a) t-SNE visualization of \textbf{classifier weights} from clients (circle) during the initial round. Different colors means different categories. Pentagram denotes the simple average of weight parameters by category. (b) \textbf{Number of excluded unlabeled samples} under different levels of heterogeneity. (c) \textbf{Test accuracy curves} of `SGD-FSSL' \emph{w/o} and \emph{w/} low-confidence samples under different ratios of correct labels. These samples could improve performance as the correctly-labeled number increases.
    }
    \label{fig:ob1-2}
\end{figure}

Each client $\mathcal{C}_k$ maintains a private partially-labeled dataset $\mathcal{D}_k$, including labeled samples $\mathcal{D}_k^\texttt{\textit{s}} {=} \{\mathbf{x}_{k,i},\mathbf{y}_{k,i}\}_{i=1}^{N_k^\texttt{\textit{s}}}$ and unlabeled samples $\mathcal{D}_k^\texttt{\textit{u}} {=} \{\mathbf{u}_{k,i}\}_{i=1}^{N_k^\texttt{\textit{u}}}$, where $N_k^\texttt{\textit{s}} \ll N_k^\texttt{\textit{u}}$. During each communication round, a subset of online clients $\mathbb{C}_M \subseteq \mathbb{C}$ is randomly selected for local training \cite{liu2025sage}. Then, FSSL methods typically aggregate the uploaded local parameters $\{\mathbf{\Theta}_m\}_{\mathcal{C}_m \in \mathbb{C}_M}$ as the global parameters $\mathbf{\Theta}_\mathcal{G} = \sum_{\mathcal{C}_m \in \mathbb{C}_M} \gamma_m \mathbf{\Theta}_m$, where the aggregation weight $\gamma_m$ of $\mathcal{C}_m$ is empirically set by the proportion of its local dataset size relative to the total samples across $M$ clients. 

\section{Preliminary Study}
\label{sec:preliminary}
In this study, we conduct some exploratory experiments for Question \ding{172} and \ding{173} in Sec.~\ref{sec:intro}.
\begin{figure*}
    \centering
    \includegraphics[width=0.9\linewidth]{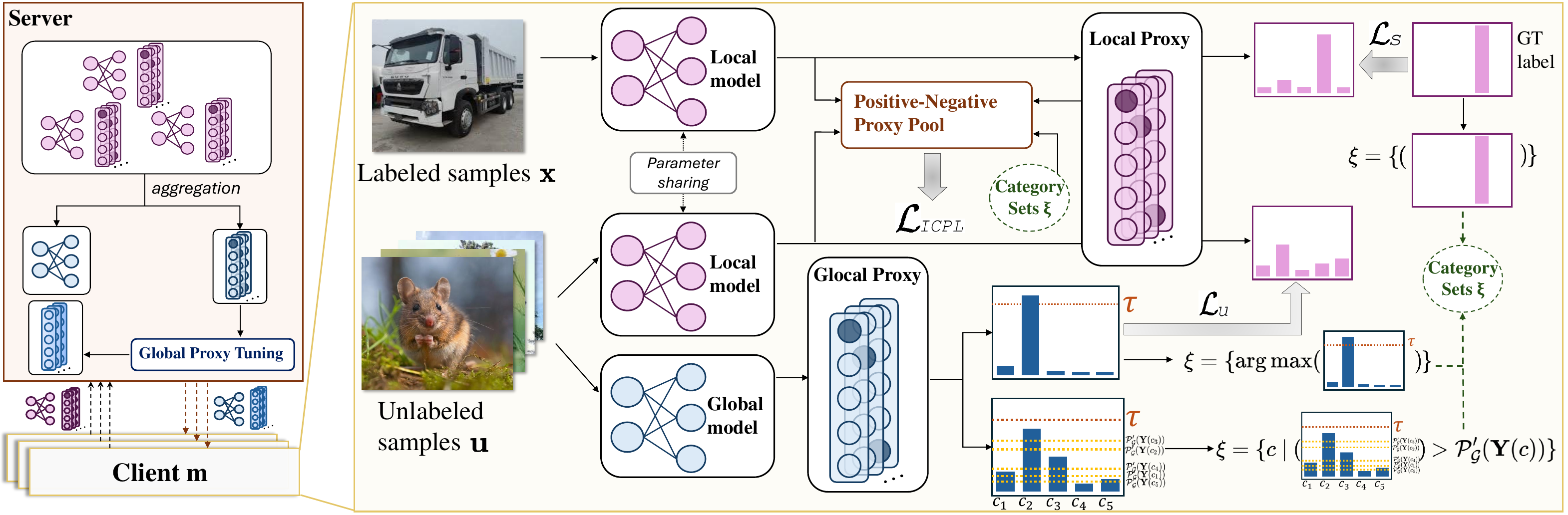}
    \caption{Overview of \textbf{ProxyFL} for Global Proxy Tuning in server-side \& Indecisive-Categories Proxy Learning in local client.}
    \label{fig:framework}
\end{figure*}
As shown in \cite{yao2022pcl}, the weight parameters of the network classifier have a certain ability in category representation. Thus, we extract the classifier weights $\{\omega_m\}_{\mathcal{C}_m\in\mathbb{C}_M},~{\omega_m\in \mathbb{R}^{C\times d}}$ from clients and slice them by class to generate a t-SNE plot, thereby visually showing the external heterogeneity to discuss Question \ding{172}. 

As shown in Fig.~\ref{fig:ob1-2}(a), directly using centroids (\emph{i.e.}, pentagrams, the averaging-based weights) may not accurately fit the global category distribution across clients. Due to external heterogeneity, some clients exhibit significant distribution discrepancy from others, \textbf{causing some points to be outliers}. So, the averaging-based method is affected by these outliers, positioning the centroid outside most points of the category cluster. We summarize as follows:
\begin{observation}
\label{ob2}
Simply averaging classifier weights is prone to skew towards outliers, thus failing to effectively capture the global category distribution across clients.
\end{observation}
For \textbf{internal heterogeneity}, most FSSL methods \cite{cho2023fedlabel,liu2025sage,zhu2024feddb} follow FixMatch \cite{sohn2020fixmatch} to keep high-confident unlabeled samples while ignoring low-confidence ones. This allows the model to heavily rely on limited easy-to-judge unlabeled samples and exacerbates internal heterogeneity. We examine internal heterogeneity in Fig.~\ref{fig:ob1-2}(b-c) and observe that:
\begin{observation}
\label{ob1}
As data heterogeneity increases, more unlabeled samples will be excluded from local training; The excluded samples have the potential to boost performance.
\end{observation}

Based on the above observations, we propose our \emph{ProxyFL} from two perspectives, \emph{i.e.}, \textbf{how to {fit} a global category distribution {robustly against outliers} (external) and how to effectively leverage low-confidence unlabeled samples (internal)}. To this end, we propose to leverage \textbf{the learnable weights of model classifier as proxy} for jointly modeling local \& global category distribution of FSSL, instead of prototypes (See Sec.~\ref{sec:empirical}). Specifically, for client $\mathcal{C}_m$, we define its proxy vectors $\mathbf{\Omega}_m$ as $\{\mathcal{\omega}_m^c\}_{c=1}^C$, \emph{i.e.}, the final FC layer of model classifier, to represent each category, where $\omega^c_m \in \mathbb{R}^d$. Our framework is shown in Fig.~\ref{fig:framework}.

\section{Methodology}
\subsection{Global Proxy Tuning}
\label{sec:gpt}
To mitigate the impact of external heterogeneity (Obs.~\ref{ob2}), we propose to model the category distribution across clients $\mathcal{P}_\mathcal{G}(\mathbf{Y})$ on the central server via learning a set of global proxies $\mathbf{\Omega}_\mathcal{G} = \{\mathcal{\omega}_\mathcal{G}^c\}_{c=1}^C$, called \underline{G}lobal \underline{P}roxy \underline{T}uning (GPT). In each federation round, the server receives model parameters from each client, thus a straight-forward idea to aggregate local classifiers is $\overline{\mathbf{\Omega}}_\mathcal{G} {=} \{\sum_{\mathcal{C}_m \in \mathbb{C}_M} \gamma_m \mathbf{\omega}_m^c\}_{c=1}^C$, where $\overline{\mathbf{\Omega}}_\mathcal{G}$ means the average operation.

However, as summarized in Obs.~\ref{ob2}, simply averaging the local proxies is prone to be affected by the outliers due to external heterogeneity. Therefore, we first initialize the global proxies $\mathbf{\Omega}_\mathcal{G}$ with $\overline{\mathbf{\Omega}}_\mathcal{G}$ and then further fine-tune $\mathbf{\Omega}_\mathcal{G}$ on the server by leveraging the off-the-shelf uploaded local proxies $\{\mathcal{\omega}_m^c\}_{c=1}^C,~\forall ~\mathcal{C}_m \in \mathbb{C}_M$. More concretely, for the global proxy ${\mathbf{\Omega}_\mathcal{G}^c}$ of category $c$, our objective is to pull it closer to all local proxies belonging to category $c$ and push it away from local proxies of other categories. The optimization objective of $\mathbf{\Omega}_\mathcal{G}$ is defined as:
\begin{align}
\min_{\mathbf{\Omega}_\mathcal{G}} \sum_{c=1}^{C} \sum_{m=1}^{M} \left [\phi(\mathbf{\Omega}_\mathcal{G}^c, \omega_m^c) -\sum_{c'=1,c' \neq c}^{C}\phi(\mathbf{\Omega}_\mathcal{G}^c, \omega_m^{c'})
\right],
\end{align}
where {$\phi(\cdot, \cdot)$} refers to the distance metric. Given the above objective, we formulate the loss function for \underline{G}lobal \underline{P}roxy \underline{T}uning (GPT) as follows:

\vspace{-0.5em}
{ \footnotesize
\begin{align}
\label{eq:loss_gpt}
\mathcal{L}_\texttt{\textit{GPT}}  =  \sum_{c=1}^{C} \sum_{m=1}^{M}-log\frac{e^{-\phi(\mathbf{\Omega}_\mathcal{G}^c,~\omega_m^c)}}{e^{-\phi(\mathbf{\Omega}_\mathcal{G}^c,~\omega_m^c)}+\sum_{c'=1,c' \neq c}^{C}e^{-\phi(\mathbf{\Omega}_\mathcal{G}^c,~\omega _m^{c'})} }.
\end{align}
}

\vspace{-0.1em}
The entire tuning process of GPT module is conducted on the server. Then, the well-tuned $\mathbf{\Omega}_\mathcal{G}$ are sent to each client together with other aggregated parameters, as the parametric initialization for the next round of local training.

\subsection{Indecisive-Categories Proxy Learning}
As noted in Obs.~\ref{ob1}, higher internal heterogeneity leads to fewer data participation. To this end, we propose \underline{I}ndecisive-\underline{C}ategories \underline{P}roxy \underline{L}earning (ICPL) to incorporate more low-confidence unlabeled samples. 

{For an unlabeled sample $\mathbf{u}_i$\footnote{Note that in this section, we drop the client subscript $m$ in data samples for symbolic simplicity.}, 
its \textbf{local feature} {$z_i$}, \textbf{local logits} {$\tilde{\mathbf{y}}_i$} and \textbf{global logits} $\overline{\mathbf{y}}_i$ via $\mathbf{\Theta}_\mathcal{G}$ can be {calculated} as:
\begin{align}
\label{eq:zyy}
 z_i & = f_{m}\left(\mathcal{T}_w(\mathbf{u}_i);\theta_{m}\right),\tilde{\mathbf{y}}_i =  h_m\left(z_i;\omega_m\right),\tilde{\mathbf{y}}_i \in \mathbb{R}^C, \\\nonumber 
& \overline{\mathbf{y}}_i = h_\mathcal{G}((f_\mathcal{G}
(\mathcal{T}_w(\mathbf{u}_i);\theta_\mathcal{G}));\mathbf{\Omega}_\mathcal{G}), \overline{\mathbf{y}}_i\in \mathbb{R}^{C}, 
\end{align}
where $\mathcal{T}_w (\cdot)$ denotes the weak augmentation.

\subsubsection{Category Set Construction} 
We first construct its category set for each sample in local client as follows: \textbf{\Rom{1}. Labeled Samples.} As shown in Fig.~\ref{fig:framework} upper right, for a labeled sample $\mathbf{x}_i$, its category set is defined by its ground-truth as $\xi_i= \{\mathbf{y}_i\}$, which is a single-element set; \textbf{\Rom{2}. High-Confidence Unlabeled Samples.} In Fig.~\ref{fig:framework} middle right, if $\max(\overline{\mathbf{y}}_i)>\tau$, $\mathbf{u}_i$ is a high-confidence sample (denoted as $\mathbf{u}_i^\texttt{\textit{hc}}$) with its pseudo-label $\hat{\mathbf{y}}_i=\arg \max(\overline{\mathbf{y}}_i)$. Here, we define its \emph{category set} $\xi_i=\{\hat{\mathbf{y}}_i\}$ as a single-element set like \textbf{\Rom{1}};
\textbf{\Rom{3}. Low-Confidence Unlabeled Samples.} In Fig.~\ref{fig:framework} lower right, if $\max(\overline{\mathbf{y}}_i)\le \tau$, $\mathbf{u}_i$ is regarded as a low-confidence sample, denoted as $\mathbf{u}_i^\texttt{\textit{lc}}$. For $\mathbf{u}_i^\texttt{\textit{lc}}$, a simple low-confident pseudo-label may affect model performance due to the potential labeling errors. Prior study \cite{chen2022vc} shows the effectiveness of assigning more than one category labels to ambiguous ROI candidates. Thus, we leverage \textbf{the several categories among which the model hesitates} to represent its possible labels for $\mathbf{u}_i^\texttt{\textit{lc}}$, defined as \emph{indecisive-categories set} ${\xi}_i$, \emph{e.g., \{mouse, hamster\}} in Fig.~\ref{fig:mpartnew}(a-b). To better determine $\xi$, we dynamically maintain a global category prior $\mathcal{P}_\mathcal{G}'(\mathbf{Y})$ to constrain the \emph{indecisive categories} for $\mathbf{u}_i^\texttt{\textit{lc}}$:
\begin{align}
\xi_i = \{c|\overline{\mathbf{y}}_i(c)>\mathcal{P}_\mathcal{G}'(\mathbf{Y}(c))\},
\end{align}
where $c\in[1,C]$. So for $\mathbf{u}^\texttt{\textit{lc}}_i$, any category $c\in[1,C]$ with corresponding global logits $\overline{\mathbf{y}}_i(c)$ exceeding $\mathcal{P}_\mathcal{G}'(\mathbf{Y}(c))$ is considered as \emph{indecisive category}. 

\noindent\textbf{Prior Distribution $\mathcal{P}_\mathcal{G}'(\mathbf{Y})$.} In each communication round $t$, we locally calculate model preference $\overline{\mathcal{P}^t_m}(\mathbf{Y})$ for category prediction, without the need of real per-class number for data privacy.
Then, they are uploaded to the server for aggregation, denoted as: 

\vspace{-0.8em}
{\scriptsize
\begin{align}
\overline{\mathcal{P}^t_m}(\mathbf{Y}) = \{\frac{\overline{N^\texttt{\textit{s,hc}}_m}(c)}{\overline{N^\texttt{\textit{s,hc}}_m}}\}_{c=1}^{C},
\overline{\mathcal{P}^t_\mathcal{G}}(\mathbf{Y}) = \{\frac{1}{M}\sum_{m=1}^{M}\overline{\mathcal{P}^t_m}(\mathbf{Y}(c))\}_{c=1}^{C},
\end{align}}

\vspace{-0.6em}
\noindent where $\mathcal{P}_\mathcal{G}'(\mathbf{Y}) \leftarrow \{\overline{\mathcal{P}^t_\mathcal{G}}(\mathbf{Y}(c))\}_{c=1}^{C}$, $\overline{N^\texttt{\textit{s,hc}}_m}(c)$ denotes the prediction counts of each category from labeled and high-confidence samples.
In ICPL, it acts as a dynamic threshold for different classes by setting a higher threshold for the majority classes and lower for the minority.

\begin{figure}
    \centering
    \includegraphics[width=1\linewidth]{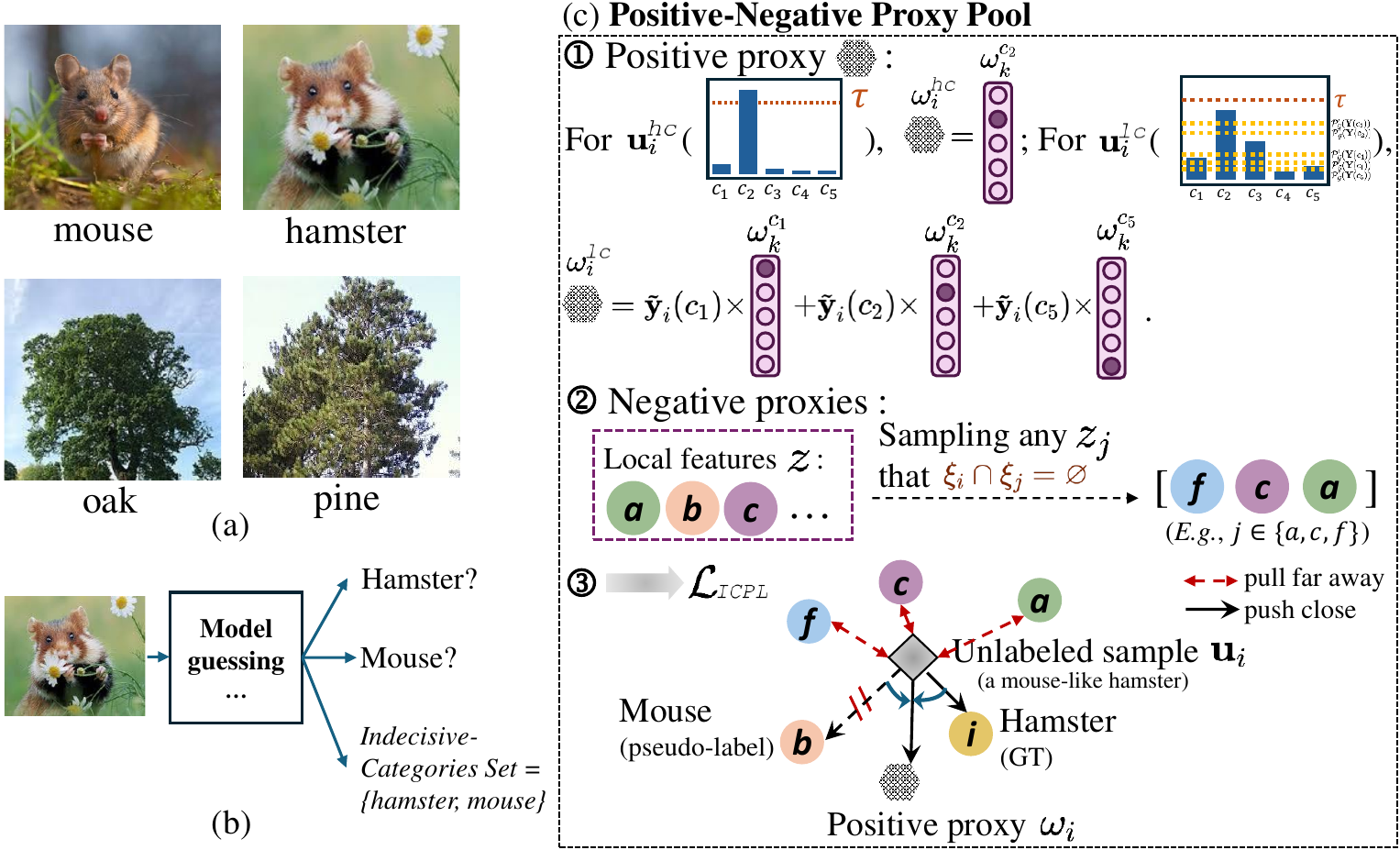}
    \caption{(a-b) Examples of the indecisive-categories. (c) Illustration of our Indecisive-Categories Proxy Learning.
    \vspace{-1em}
    }
    \label{fig:mpartnew}
\end{figure}

\subsubsection{Positive-Negative Proxy Pool}
Based on the \textbf{category set} of each sample, we build \textbf{the relationship among all types of samples} and train them by Contrastive Learning (CL) to mitigate internal heterogeneity.
To this end, we construct a `Positive-Negative Proxy Pool' as follows (shown in Fig.~\ref{fig:mpartnew}(c)):
\begin{remark}
\label{remark:pool}
\textup{\textbf{(Positive-Negative Proxy Pool)}} For a batch $\mathcal{B}$ from $\mathcal{D}_m$, we construct the Positive-Negative Proxy Pool without excluding any of the samples. \textup{\textbf{1) Positive Proxy.}} The positive proxy of a high-confidence sample $\mathbf{u}_i^\texttt{\textit{hc}}$ and a low-confidence sample $\mathbf{u}_i^\texttt{\textit{lc}}$ are respectively defined as:
\begin{align}
\label{eq:posproxy}
\omega_i^\texttt{\textit{hc}} = \omega_k^{\hat{\mathbf{y}_i}},~\omega_i^\texttt{\textit{lc}} = \sum_{\forall {c’}\in\xi_i} \tilde{\mathbf{y}}_i({c’}) \times\omega _k^{c’}, 
\end{align}

\vspace{-0.5em}
\noindent where for $\mathbf{u}_i^\texttt{\textit{hc}}$, $\hat{\mathbf{y}_i}$ denotes argmax-derived category from its pseudo-label; for $\mathbf{u}_i^\texttt{\textit{lc}}$, $\omega_i^\texttt{\textit{lc}}$ is designed to the weighted sum of proxy weights, based on the categories $c'$ from $\xi_i$. 

\noindent\textup{\textbf{2) Negative Proxies Set.}} For an unlabeled sample $\mathbf{u}_i$ from batch $\mathcal{B}$, any other sample $j$ in $\mathcal{B}$ (including $\mathbf{x}_j$/$\mathbf{u}_j^\texttt{\textit{hc}}$/$\mathbf{u}_j^\texttt{\textit{lc}}$) will be considered as one of the negative-proxy candidates as long as its category set $\xi_j$ does not overlap with $\xi_i$, i.e.,
\vspace{-0.3em}
\begin{align}
\label{eq:negproxy}
\mathcal{R}_i & = \left\{ z_j ~\middle|~ \left [   z_j = f_m(\mathbf{x}_j;\theta _m) \vee  z_j = \right. \right. \\ \nonumber
& \quad \left. \left. f_m(\mathbf{u}_j;\theta _m)\right ] \land \xi_j \cap \xi_i=\varnothing  \right\}.
\end{align}

\end{remark}
\begin{table*}[]
    \centering
    \caption{ Experimental results on CIFAR-10, CIFAR-100, SVHN and CINIC-10 under 10\% label. Bold text indicates the best result, and the last row presents the improvement of ProxyFL over the second best method.}
    \renewcommand{\arraystretch}{1}
    \resizebox{0.98\linewidth}{!}{
    \begin{tabular}{c|ccc|ccc|ccc|ccc}
    \hline\hline
         \multirow{2}{*}{\textbf{Methods}} & \multicolumn{3}{c}{CIFAR-10} & \multicolumn{3}{c}{CIFAR-100} & \multicolumn{3}{c}{SVHN} & \multicolumn{3}{c}{CINIC-10}\\
         & $\alpha=0.1$ & $\alpha=0.5$ & $\alpha=1$ & $\alpha=0.1$ & $\alpha=0.5$ & $\alpha=1$ & $\alpha=0.1$ & $\alpha=0.5$ & $\alpha=1$ & $\alpha=0.1$ & $\alpha=0.5$ & $\alpha=1$\\
    \hline\hline
        \multicolumn{1}{c}{\textbf{FL Methods}} & \multicolumn{12}{c}{} \\
    \hline\hline
        FedAvg \cite{mcmahan2017fedavg} & 69.60 & 68.88 & 69.39 & 34.08 & 33.21 & 35.31 & 82.40 & 83.40 & 78.60 & 57.17 & 60.09 & 61.54 \\
     
        FedProx \cite{li2020fedprox} & 68.58 & 69.53 & 68.00 & 34.20 & 34.07 & 34.88 & 81.67 & 83.77 & 83.77 & 58.05 & 60.71 & 62.82 \\

        FedAvg-SL & 90.46 & 91.24 & 91.32 & 67.98 & 68.83 & 69.10 & 94.11 & 94.41 & 94.40 & 77.82 & 80.42 & 81.29 \\
    \hline\hline
        \multicolumn{1}{c}{\textbf{FL+SSL Methods}} & \multicolumn{12}{c}{} \\
    \hline\hline
        FixMatch-LPL & 82.98 & 84.36 & 84.69 & 49.32 & 49.67 & 49.55 & 89.68 & 91.33 & 91.91 & 68.02 & 70.67 & 72.69 \\

        FixMatch-GPL & 84.56 & 86.05 & 86.66 & 48.96 & 51.80 & 52.19 & 90.50 & 91.94 & 92.31 & 71.67 & 73.26 & 74.80 \\

        FedProx+FixMatch & 84.60 & 85.49 & 86.95 & 48.42 & 48.51 & 49.33 & 90.46 & 91.36 & 91.25 & 68.62 & 70.67 & 72.69 \\

        FedAvg+FlexMatch & 84.21 & 86.00 & 86.57 & 49.91 & 51.39 & 51.79 & 52.58 & 55.59 & 60.50 & 69.20 & 71.87 & 73.42 \\
    \hline\hline
        \multicolumn{1}{c}{\textbf{FSSL Methods}}& \multicolumn{12}{c}{} \\
    \hline\hline
        FedMatch \cite{jeong2020fedmatch} & 75.35 & 77.86 & 78.00 & 32.23 & 31.49 & 35.75 & 88.63 & 89.20 & 89.23 & 51.94 & 56.27 & 70.22 \\

        FedLabel \cite{cho2023fedlabel} & 62.85 & 79.46 & 79.17 & 50.88 & 52.21 & 52.38 & 89.31 & 91.51 & 91.16 & 67.64 & 70.56 & 72.80 \\

        FedLoke \cite{zhang2023fedloke} & 83.32 & 82.22 & 81.87 & 39.29 & 40.46 & 39.96 & 89.94 & 90.00 & 89.45 & 59.03 & 61.60 & 63.21 \\

        FedDure \cite{bai2024feddure} & 84.60 & 85.88 & 87.34 & 48.27 & 51.09 & 50.79 & 92.87 & 93.49 & 94.19 & 70.86 & 73.37 & 74.89 \\

        FedDB \cite{zhu2024feddb} & 83.99 & 85.28 & 87.49 & 48.43 & 50.11 & 51.55 & 92.56 & 93.00 & 93.14 & 69.44 & 72.60 & 73.61 \\

        SAGE \cite{liu2025sage} & 87.05 & 88.05 & 89.08 & 54.18 & 55.82 & 56.06 & 93.85 & 94.27 & 94.65 & 74.59 & 75.74 & 76.68 \\
        
        \multirow{2}{*}{\textbf{ProxyFL (ours)}} & \cellcolor[gray]{0.9}\textbf{88.56} & \cellcolor[gray]{0.9}\textbf{90.00} & \cellcolor[gray]{0.9}\textbf{89.96} & \cellcolor[gray]{0.9}\textbf{57.50} & \cellcolor[gray]{0.9}\textbf{58.75} & \cellcolor[gray]{0.9}\textbf{58.24} & \cellcolor[gray]{0.9}\textbf{95.09} & \cellcolor[gray]{0.9}\textbf{95.18} & \cellcolor[gray]{0.9}\textbf{95.26} & \cellcolor[gray]{0.9}\textbf{77.98} & \cellcolor[gray]{0.9}\textbf{78.96} & \cellcolor[gray]{0.9}\textbf{79.59} \\ 
        & \textcolor{RoyalBlue}{$\uparrow$ 1.51} & \textcolor{RoyalBlue}{$\uparrow$ 1.95} & \textcolor{RoyalBlue}{$\uparrow$ 0.88} & \textcolor{RoyalBlue}{$\uparrow$ 3.32} & \textcolor{RoyalBlue}{$\uparrow$ 2.93} & \textcolor{RoyalBlue}{$\uparrow$ 2.18} & \textcolor{RoyalBlue}{$\uparrow$ 1.24} & \textcolor{RoyalBlue}{$\uparrow$ 0.91} & \textcolor{RoyalBlue}{$\uparrow$ 0.61} & \textcolor{RoyalBlue}{$\uparrow$ 3.39} & \textcolor{RoyalBlue}{$\uparrow$ 3.22} & \textcolor{RoyalBlue}{$\uparrow$ 2.91} \\
    \hline\hline
    \end{tabular}
    }
    \label{tab:comparison}
\end{table*}
According to Eq.~\ref{eq:posproxy}-\ref{eq:negproxy}, we ensure more data participation into training while reducing potential errors of pseudo-labels. The objective of ICPL can be formulated as:
\begin{align}
\label{eq:icpl}
\mathcal{L}_\texttt{\textit{ICPL}}  = -\left[\frac{1}{|\mathcal{B}^{\texttt{\textit{u}},\texttt{\textit{hc}}}|}\sum_{i=1}^{|\mathcal{B}^{\texttt{\textit{u}},\texttt{\textit{hc}}}|}\log \frac{e^{z_i~\cdot~\omega_i^\texttt{\textit{hc}}}}{e^{z_i~\cdot~\omega_i^\texttt{\textit{hc}}}+\sum_{z_j\in\mathcal{R}_i} e^{z_i~\cdot ~z_j}}   \right. \\\nonumber \left. +\frac{1}{|\mathcal{B}^{\texttt{\textit{u}},\texttt{\textit{lc}}}|}\sum_{i=1}^{|\mathcal{B}^{\texttt{\textit{u}},\texttt{\textit{lc}}}|}\log \frac{e^{z_i~\cdot~\omega_i^\texttt{\textit{lc}}}}{e^{z_i~\cdot~\omega_i^\texttt{\textit{lc}}}+\sum_{z_j\in\mathcal{R}_i} e^{z_i~\cdot ~z_j}} \right]. 
\end{align}
where $|\mathcal{B}^{\texttt{\textit{u}},\texttt{\textit{hc}}}|$, $|\mathcal{B}^{\texttt{\textit{u}},\texttt{\textit{lc}}}|$ denotes the number of high-confidence samples and low-confidence samples in batch $\mathcal{B}^\texttt{\textit{u}}$, respectively. 
\subsection{Loss Functions}
In local training, we follow previous studies \cite{li2023cbafed} to assign \emph{ground-truth} $\mathbf{y}$ for labeled data and \emph{pseudo-label} $\hat{\mathbf{y}}$ for high-confidence unlabeled data, respectively. Following SAGE \cite{liu2025sage}, the local losses are:
\vspace{-0.7em}
\begin{align}
\label{eq:kl_sup_loss_custom}
    \mathcal{L}_\texttt{\textit{u}} &= \frac{1}{|\mathcal{B}^{\texttt{\textit{u}},\texttt{\textit{hc}}}|}
    \sum_{i=1}^{|\mathcal{B}^{\texttt{\textit{u}},\texttt{\textit{hc}}}|}
    \mathbf{KL}(h_m(f_{m}(\mathcal{T}_s(\mathbf{u}_i); \theta_{m});\omega_m) \parallel \hat{\mathbf{y}}_i), \nonumber \\[-6pt]  
    \mathcal{L}_\texttt{\textit{s}} &= \frac{1}{|\mathcal{B}^\texttt{\textit{s}}|}
    \sum_{i=1}^{|\mathcal{B}^\texttt{\textit{s}}|} \mathcal{L}_\texttt{\textit{CE}}(h_m(f_{m}(\mathbf{x}_i; \theta_{m});\omega_m),\mathbf{y}_i),
\end{align}
where $\mathbf{KL}$ denotes Kullback-Leibler divergence loss, $\mathcal{T}_s(\cdot)$ denotes the strong-augmentation, $\mathcal{L}_\textit{\texttt{CE}}$ denotes cross-entropy loss. Following these, 
our ProxyFL locally incorporate low-confidence unlabeled samples 
via $\mathcal{L}_\texttt{\textit{ICPL}}$ and improve the global category distribution via $\mathcal{L}_\texttt{\textit{GPT}}$ on the server. Thus, our final total objective is:
\begin{align}
\mathcal{L}=\underbrace{\mathcal{L}_\texttt{\textit{s}}+\alpha \mathcal{L}_\texttt{\textit{u}}+\beta\mathcal{L}_\texttt{\textit{ICPL}}}_{\textbf{local}} +\underbrace{ \mathcal{L}_\texttt{\textit{GPT}}}_{\textbf{global}}, 
\end{align}
where $\alpha,\beta$ are empirically set to $1$. The pseudo-code of ProxyFL is shown in \textbf{Appendix.~\ref{appendix:algo}}. We also provide some \textbf{theoretical proofs} of ProxyFL in \textbf{Appendix.~\ref{appendix:proof}}.

\section{Experiments}
\label{sec:experiments}
\subsection{Experimental Setup}
\label{sec:setup}
\noindent\textbf{Datasets.} We strictly follow the FSSL experimental setting of SAGE \cite{liu2025sage} which is evaluated on the CIFAR10, CIFAR-100, SVHN, and CINIC-10 datasets \cite{darlow2018cinic,krizhevsky2009learning,netzer2011reading}. We partition the labeled and unlabeled samples per category with label proportions of $10\%$ and $20\%$ for each dataset. Following previous FSSL works \cite{zhu2024feddb,bai2024feddure,cho2023fedlabel}, we simulate internal and external heterogeneity by sampling labeled \& unlabeled data from three levels of Dirichlet distribution $Dir(\alpha)$: $\alpha = \{0.1,0.5,1\}$ and allocating them to local clients. The smaller $\alpha$, the higher FL data heterogeneity. We visualize the specific data distribution in Fig.~\ref{fig:exp}(a).

\noindent\textbf{Implementation details} Following SAGE \cite{liu2025sage}, we configure 20 clients for all settings, with 8 clients randomly sampled each round to participate in the federated training. ResNet-8 \cite{he2016resnet} serves as the local backbone, with the number of local epochs set to 5, local learning rate set to 0.1 and the confidence threshold $\tau$ for pseudo-labeling set to 0.95. For global proxy tuning process, the learning rate is 0.005, and the number of tuning epochs is set to 10 for CIFAR-100 and 100 for the other datasets. Unless otherwise specified, the experimental setup of ProxyFL is consistent with SAGE (More details in Appendix.~\ref{appendix:add_exp}).

\begin{table*}[t]
    \centering
    \caption{Module ablation studies on GPT and ICPL of our method.}
    \renewcommand{\arraystretch}{1}
    \resizebox{0.9\linewidth}{!}{
    \begin{tabular}{cc|ccc|ccc|ccc|ccc}
    \hline\hline
     \multirow{2}{*}{GPT} & \multirow{2}{*}{ICPL} & \multicolumn{3}{c}{CIFAR10} & \multicolumn{3}{c}{CIFAR100} & \multicolumn{3}{c}{SVHN} & \multicolumn{3}{c}{CINIC10}\\
    & & $\alpha=0.1$ & $\alpha=0.5$ & $\alpha=1$ & $\alpha=0.1$ & $\alpha=0.5$ & $\alpha=1$ & $\alpha=0.1$ & $\alpha=0.5$ & $\alpha=1$ & $\alpha=0.1$ & $\alpha=0.5$ & $\alpha=1$\\
    \hline\hline
     & & 84.56 & 86.05 & 86.66 & 48.96 & 51.80 & 52.19 & 90.50 & 91.94 & 92.31 & 71.67 & 73.26 & 74.80 \\
    \checkmark & & 87.59 & 89.23 & 89.71 &54.86 & 56.58 & 57.09 &94.29 & 94.49 & 94.53 & 77.15 & 79.03 & 79.31 \\
    & \checkmark & 87.81 & 89.58 & 89.66 & 57.21 & 57.98 & 57.74 & 94.82 & 94.69 & 95.15 & 77.80 & 78.04 & 78.57 \\
    \checkmark & \checkmark & \textbf{88.56} & \textbf{90.00} & \textbf{89.96} & \textbf{57.50} & \textbf{58.75} & \textbf{58.24} & \textbf{95.09} & \textbf{95.18} & \textbf{95.26} & \textbf{77.98} & \textbf{78.96} & \textbf{79.59} \\
    \hline\hline
    \end{tabular}
    }
    \label{tab:ablation}
\end{table*}
\begin{figure*}
    \centering
    \includegraphics[width=0.95\linewidth]{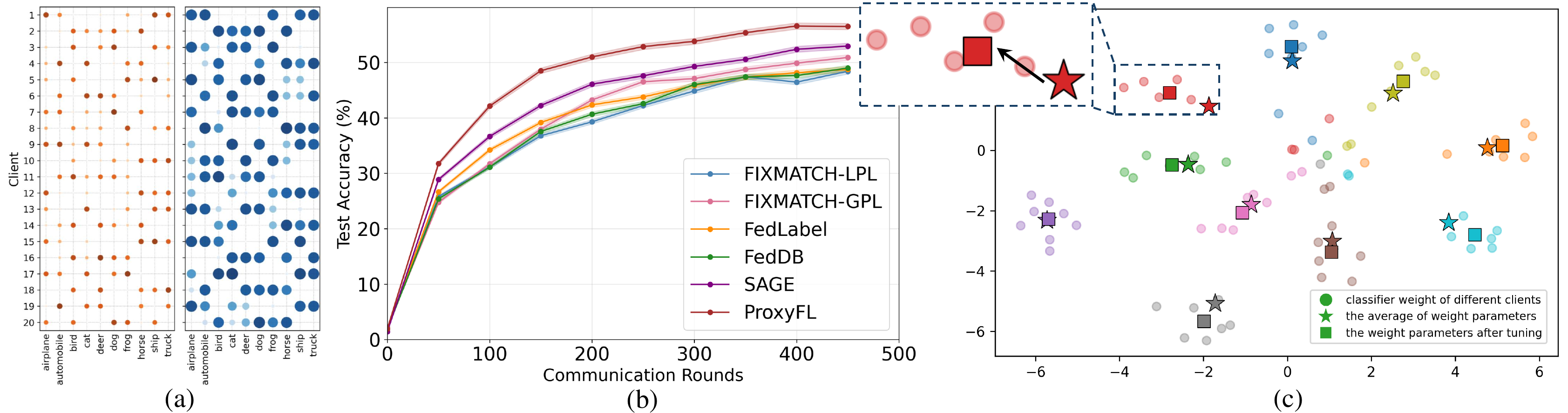}
    \caption{(a) Distribution of labeled and unlabeled data across clients under $\alpha=0.1$ taking CIFAR-10 as an example. (b) Convergence curves of ProxyFL and other baselines on CIFAR-100 with $\alpha = 0.1$. (c) Distribution of global category proxies before-and after-tuning visualized in a t-SNE plot.}
    \label{fig:exp}
\end{figure*}

\subsection{Performance Comparison}
\label{performance_comparison}
Tab.~\ref{tab:comparison} reports the overall results of our ProxyFL and other state-of-the-art methods across multiple datasets under different Non-IID (\textbf{Non}-\textbf{I}ndependent and \textbf{I}dentical \textbf{D}istribution) scenarios with $10\%$ label. We compare the following methods in our experiments like SAGE \cite{liu2025sage}: \textbf{\Rom{1}. FL methods.} For FedAvg \cite{mcmahan2017fedavg} and FedProx \cite{li2020fedprox}, all clients are trained via supervised federated learning only on partial labeled data; For FedAvg-SL, local data are all fully-labeled data, which denotes the ideal upper-bound based on standard fully-supervised FedAvg. \textbf{\Rom{2}. Vanilla combinations (FL + SSL methods).} Here, each method denotes a simple combination of SSL methods and FL methods. Note that FixMatch-LPL and FixMatch-GPL are both FixMatch-based frameworks, but pseudo-labels (PL) are derived from different models, \emph{i.e.}, local model for LPL and global model for GPL, respectively. \textbf{\Rom{3}. FSSL methods.} We compare ProxyFL with previous state-of-the-art FSSL methods, including FedMatch \cite{jeong2020fedmatch}, FedLabel \cite{cho2023fedlabel}, FedLoke \cite{zhang2023fedloke}, FedDure \cite{bai2024feddure}, FedDB \cite{zhu2024feddb} and SAGE \cite{liu2025sage}.

In Tab.~\ref{tab:comparison}, ProxyFL achieves state-of-the-art performances on multiple datasets with significant improvements under different levels of data heterogeneity $\alpha$. \textbf{To the best of our knowledge, we are the first in FSSL to propose category proxy for mitigating both internal and external heterogeneity.} Notably, our ProxyFL even achieves comparable performance to that of FedAvg-SL on SVHN dataset and CINIC-10 when $\alpha=0.1$. We attribute this improvement to the generalization of the enhanced category distribution from our proxy-guided FSSL framework. 

\subsection{Empirical Analysis}
\label{sec:empirical}

In this section, we conduct an in-depth investigation to assess the contributions of our GPT and ICPL in ProxyFL. 

\noindent\textbf{1) Effectiveness of modules} We first assess the contributions of GPT and ICPL through ablation experiments and set FedAvg+FixMatch-GPL as our baseline model. We conduct ablation studies under different levels of heterogeneity $\alpha=\{0.1,~0.5,~1\}$ to validate the effectiveness of each module. As shown in Tab.~\ref{tab:ablation}, each module could individually enhance model performance and their combination of GPT \& ICPL achieves the best results\vspace{0.5em}.

\noindent\textbf{2) Convergence analysis} Fig.~\ref{fig:exp}(b) and Tab.~\ref{tab:convergence_rate} demonstrate that ProxyFL substantially improves the convergence speed and test accuracy on the CIFAR-100 dataset with $\alpha=0.1$. ProxyFL outperforms baseline and current FSSL methods, achieving better performance with greater communication efficiency. Most existing FSSL methods \cite{jeong2020fedmatch,cho2023fedlabel,zhu2024feddb} only retain high-confidence samples for training while discarding the low-confidence ones, leading to slower model convergence due to fewer training samples under non-IID scenarios. In contrast, ProxyFL cautiously incorporates low-confidence unlabeled samples into training and leverages category proxies to address both internal and external heterogeneity, accelerating model convergence especially during the early training stages (Tab.~\ref{tab:convergence_rate}). We also provide some \textbf{theoretical analysis \& proofs} in Appendix.~\ref{appendix:proof}\vspace{0.5em}.

\begin{table}
    \centering
    \scriptsize
    \caption{Comparison of convergence rates between ProxyFL and other baselines with $\alpha = 0.1$ on CIFAR-100.}
    \renewcommand{\arraystretch}{1}
    \setlength{\tabcolsep}{0.8pt} 
    \begin{tabular}{ccccccc}
    \hline\hline
        Acc. & \multicolumn{2}{c}{30\%} & \multicolumn{2}{c}{40\%} & \multicolumn{2}{c}{50\%} \\
        Method & Round$\downarrow$ & Speedup$\uparrow$ & Round$\downarrow$ & Speedup$\uparrow$ & Round$\downarrow$ & Speedup$\uparrow$ \\
    \hline\hline
        LPL & 119 & $\times$1.00 & 242 & $\times$1.00 & 562 & $\times$1.00 \\
        GPL & 114 & $\times$1.04 & 226 & $\times$1.07 & 524 & $\times$1.07 \\
        FedLabel & 94 & $\times$1.27 & 175 & $\times$1.38 & 429 & $\times$1.31 \\
        FedDB & 103 & $\times$1.16 & 206 & $\times$1.17 & - & - \\
        FedDure & 114 & $\times$1.04 & 234 & $\times$1.03 & 542 & $\times$1.04 \\
        SAGE & 60 & $\times$1.98 & 124 & $\times$1.95 & 267 & $\times$2.10 \\
        
        \textbf{ProxyFL} & \cellcolor[gray]{0.9}\textbf{45} & $\times$\cellcolor[gray]{0.9}\textbf{2.64} & \cellcolor[gray]{0.9}\textbf{89} & $\times$\cellcolor[gray]{0.9}\textbf{2.72} & \cellcolor[gray]{0.9}\textbf{177} & $\times$\cellcolor[gray]{0.9}\textbf{3.18} \\
    \hline\hline
    \end{tabular}
    \label{tab:convergence_rate}
\end{table}

\noindent\textbf{3) Analysis of Global Proxy Tuning} 

\textbf{\Rom{1}. Global Category Distribution} We explore the effect of our GPT module by visualizing the proxy distribution across clients in a t-SNE plot. As observed in Fig.~\ref{fig:exp}(c), the squares (\emph{the proxies after tuning}) fit more accurately the proxy distribution across clients than the pentagram centroids (\emph{the directly-averaging proxies}), showing better robustness to outliers. By explicitly defining an proxy optimization objective, our method can effectively fit the global distribution of per-category proxies across clients. 

\textbf{\Rom{2}. Low-Overhead Proxy Tuning} The computational complexity of our tuning process can be formulated as $O_\textit{\texttt{GPT}} = O(Q{\times} M {\times} C^2 {\times} d)$, where $Q$, $M$, $C$, and $d$ denote the number of tuning epochs, number of clients, category number, and proxy dimension, respectively. Compared to high-dimensional data/features, our tuning is conducted on lightweight low-dimensional proxy vectors, which results in much lower overheads. Taking CIFAR100 as an example, $O_\textit{\texttt{GPT}} \approx 0.4$ GFLOPs, which corresponds to the overhead of inferring just one image via ResNet (\emph{i.e.}, $0.5 \sim 1$ GFLOPs). Thus, the server-side tuning overheads could be negligible.

\begin{table}
    \centering
    \caption{Ablation of design choices between ICPL and other methods under $\alpha = 0.1$.}
    \renewcommand{\arraystretch}{1}
    \resizebox{0.95\linewidth}{!}{
    \begin{tabular}{ccccccc}
    \hline\hline
        Methods           & CIFAR-10 & CIFAR-100 & SVHN  & CINIC-10 \\
    \hline\hline
        FedAvg-SL & 90.46   & 67.98   & 94.11 & 77.82 \\ 
    \hline\hline
        GPL      & 84.56   & 48.96   & 90.50 & 71.67 \\ 
        GPL-ALL  & 85.38   & 50.34   & 93.31 & 75.83 \\
        LPL     & 82.98   & 49.32   & 89.68 & 68.02 \\ 
        LPL-ALL & 87.18   & 53.28   & 83.99 & 73.39 \\ 
    \hline\hline
        ICPL-Top1 & 87.13 & 55.66 & 94.56 & 77.01\\
        ICPL-Top5 & 87.77 & 56.58 & 94.71 & 77.65\\
        \textbf{ICPL}-$\mathcal{P}_\mathcal{G}'(\mathbf{Y})$ & \cellcolor[gray]{0.9}\textbf{87.81} & \cellcolor[gray]{0.9}\textbf{57.21} & \cellcolor[gray]{0.9}\textbf{94.82} & \cellcolor[gray]{0.9}\textbf{77.80} \\
    \hline\hline
    \end{tabular}
    }
    \label{tab:icpl_ablation}
\end{table}

\begin{figure}
    \centering
    \includegraphics[width=0.85\linewidth]{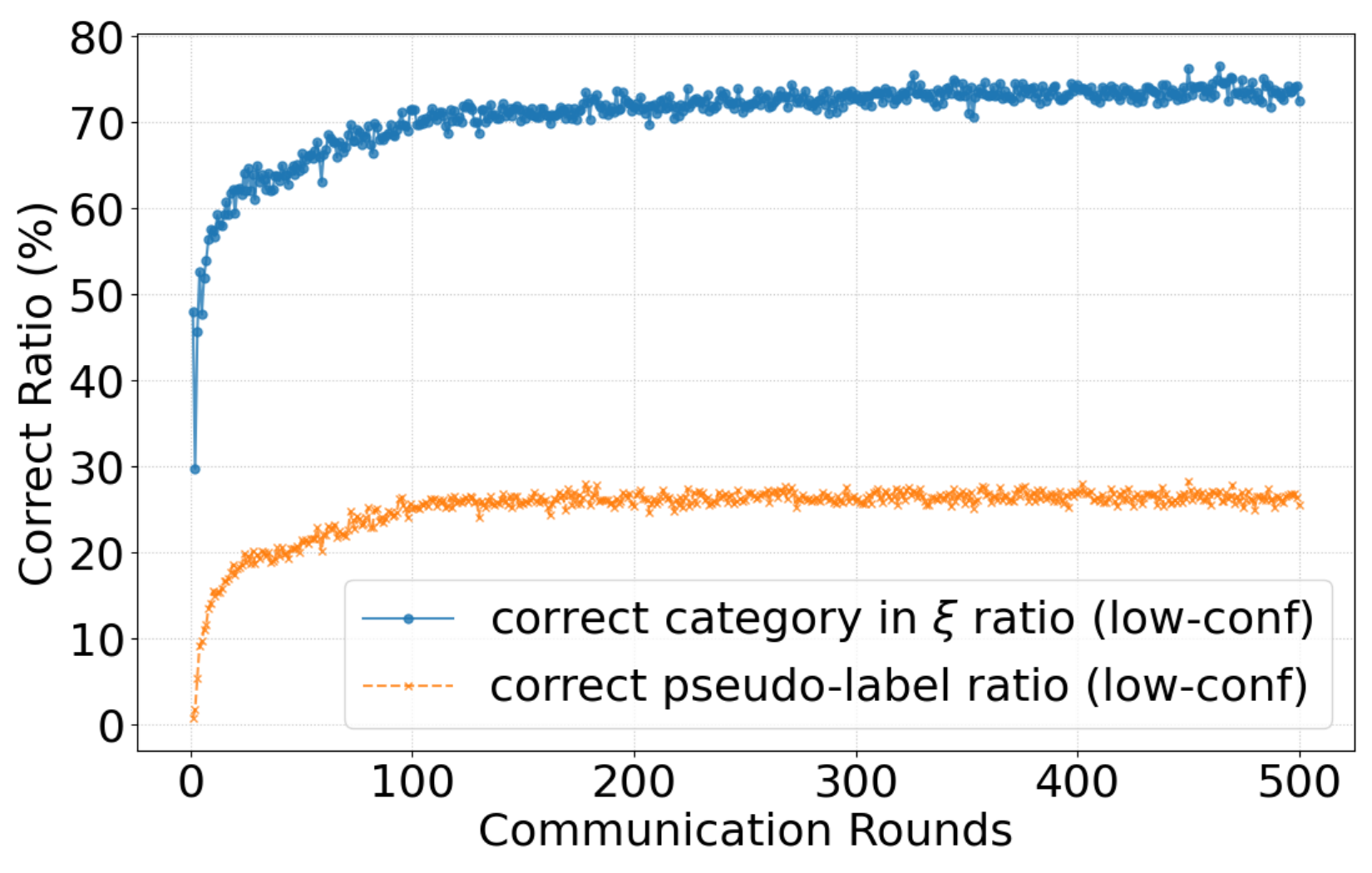}
    \caption{Ratios of `Correct category in $\xi$' vs. `Correct pseudo-label' for low-confidence samples.\vspace{-1em}}
    \label{fig:low_conf_in}
\end{figure}

\noindent\textbf{4) Analysis of Indecisive-Categories Proxy Learning} 

\textbf{\Rom{1}. Design of including low-confidence samples} An intuitive idea to include low-confidence samples $\mathbf{u}^\textit{\texttt{lc}}$ is to directly assign pseudo-labels for them like high-confidence samples, abbreviated as LPL-ALL and GPL-ALL. FedAvg-SL, the standard fully-labeled FedAvg, serves as an upperbound with correct labels. As shown in Tab.~\ref{tab:icpl_ablation} upper, in most cases, directly including $\mathbf{u}^\textit{\texttt{lc}}$ (\emph{w/}-ALL) could bring slight improvements compared to simply-discarding (\emph{w/o}-ALL), suggesting that $\mathbf{u}^\textit{\texttt{lc}}$ contain some valuable information and simply discarding them may exclude some correctly-labeled samples from training; But, directly including them sometimes leads to performance degradation, \emph{e.g.}, LPL \& LPL-ALL on SVHN. Compared to discarding or directly including $\mathbf{u}^\textit{\texttt{lc}}$, our proposed ICPL module achieves better performance across all datasets by more accurately constructing the relationships between samples in the positive-negative pool of ICPL. Moreover, ICPL even reaches the performance of FedAvg-SL on certain datasets. 

\textbf{\Rom{2}. Design of indecisive-categories set $\xi$}
We discuss the design of \emph{indecisive-categories set} $\xi$ for low-confidence samples $\mathbf{u}^\texttt{\textit{lc}}$. We compare our strategy of the prior $\mathcal{P}_\mathcal{G}'(\mathbf{Y})$ for $\xi$ with other designs: Select the multiple categories from Top-1 or Top-5 confidence scores. As shown in Tab.~\ref{tab:icpl_ablation} lower, our method consistently yields better performance than other designs, validating the effectiveness of $\mathcal{P}_\mathcal{G}'(\mathbf{Y})$ that sets different thresholds for different categories.  

\textbf{\Rom{3}. Indecisive-categories set $\xi$ vs. pseudo-label} 
We further validate the superiority of $\xi$ over pseudo-labels for low-confidence unlabeled samples $\mathbf{u}^\textit{\texttt{lc}}$. Intuitively, a set with multiple categories has higher probability to cover correct category than a single pseudo-label.
To this end, we plot 'correct category in $\xi$' and 'correct pseudo-label' ratios on $\mathbf{u}^\textit{\texttt{lc}}$. As shown in Fig.~\ref{fig:low_conf_in}, the recall of $\xi$ remains at a considerably high level, significantly outperforming the accuracy of pseudo-labeling even throughout the training process. Although this set-based supervision is not as precise as the ground-truth, it effectively introduce more available $\mathbf{u}^\textit{\texttt{lc}}$ rather than compromise on wrong pseudo-labels\vspace{0.5em}.

\noindent\textbf{5) Proxy vs. Prototypes} 
Previous FL approaches \cite{tan2022fedproto,huang2023rethinking} often employ prototypes to refine the local and global category distribution. However, high-dimensional features (prototypes) poses the risk of reversely reconstruction \cite{dosovitskiy2016inverting,melis2019exploiting}. In contrast, proxies serve as the intrinsic model parameters without raising privacy concerns \& extra overheads. We compare our method with `FedProto \cite{tan2022fedproto} + FSSL' with different strategies and Tab.~\ref{tab:proto}
validate our superiority.

\begin{table}[]
    \centering
    \caption{Ablation of our ProxyFL vs. Prototypes with $\alpha = 0.1$. 'ALL' means generating prototypes with low-confidence samples, and 'Labeled' with only labeled samples.}
    \renewcommand{\arraystretch}{1}
    \resizebox{0.95\linewidth}{!}{
    \begin{tabular}{ccccccc}
    \hline\hline
        Methods           & CIFAR-10 & CIFAR-100 & SVHN  & CINIC-10 \\
    \hline\hline
        PROTO      & 85.89   & 49.56   & 93.48 & 72.53 \\ 
        PROTO-ALL  & 86.56   & 54.69   & 19.59 & 73.16 \\
        PROTO-Labeled   & 86.55   & 48.83   & 93.60 & 72.57 \\ 
\textbf{ICPL}-$\mathcal{P}_\mathcal{G}'(\mathbf{Y})$
        & \cellcolor[gray]{0.9}\textbf{87.81} & \cellcolor[gray]{0.9}\textbf{57.21} & \cellcolor[gray]{0.9}\textbf{94.82} & \cellcolor[gray]{0.9}\textbf{77.80} \\
    \hline\hline
    \end{tabular}
    }\vspace{-1em}
    \label{tab:proto}
\end{table}

\section{Conclusion}
Our paper proposes a new Federated Semi-Supervised Learning (FSSL) method called ProxyFL, leveraging a unified proxy to simultaneously mitigate external and internal heterogeneity. We model the category distribution both locally and globally. Firstly, we define a global tuning objective to optimize the global category distribution across clients, mitigating distribution shift from external heterogeneity. Secondly, we incorporate low-confidence unlabeled samples via our dynamic \emph{indecisive-categories} proxy learning mechanism to mitigate internal heterogeneity. Extensive experiments show that ProxyFL can significantly boost the performance and convergence of the FSSL model.

\section{Acknowledgement}
This work was supported by the National Natural Science Foundation of China (Grant No. 62471182), Science and Technology Commission of Shanghai Municipality Basic Research Program (Grant No. 25JD1401300), Shanghai Rising-Star Program (Grant No. 24QA2702100), and the Science and Technology Commission of Shanghai Municipality (Grant No. 22DZ2229004).

{
    \small
    \bibliographystyle{ieeenat_fullname}
    \bibliography{main}
}

\newpage
\appendix

\section{Definitions of Data Heterogeneity in FSSL}
\label{appendix:def_of_dh}
We define data heterogeneity in FSSL as follows:
\begin{definition}
\label{def:ext}
\textup{\textbf{(External heterogeneity in FSSL)}} External heterogeneity refers to the distribution discrepancy between $\mathcal{D}_k$ across different clients $\{\mathcal{C}_1, \ldots, \mathcal{C}_K\}$, \emph{i.e.}, for any two different clients $\mathcal{C}_{k_1}$ and $\mathcal{C}_{k_2}$, $\mathcal{P}_{k_1}(\mathbf{Y}) {\neq} \mathcal{P}_{k_2}(\mathbf{Y})$.
\end{definition}
\begin{definition}
\label{def:in}
\textup{\textbf{(Internal heterogeneity in FSSL)}} Internal heterogeneity exists within local clients, embodied in: 1) class imbalance, arising from unequal sample sizes among {different} categories within $\mathcal{C}_k$, \emph{i.e.}, for any two categories $c_1$ and $c_2$, $\mathcal{P}_k(\mathbf{Y}({c_1})) {\neq} \mathcal{P}_k(\mathbf{Y}({c_2}))$; 2) distribution imbalance between labeled and unlabeled data, i.e., $\mathcal{P}_k^\texttt{\textit{s}}(\mathbf{Y}) {\neq} \mathcal{P}_k^\texttt{\textit{u}}(\mathbf{Y})$.
\end{definition}

\section{Pseudo-Code}
\label{appendix:algo}
The pseudo-code of ProxyFL is shown in Algorithm.~\ref{alg:proxyfl}.

{ 
\setlength{\algomargin}{0em} 
\begin{algorithm}[h]
\small
\caption{Proxy-Guided FSSL (ProxyFL)}
\label{alg:proxyfl}
\KwIn{
Federation system $\mathbb{C}$,
communication round $T$,
local learning rate $\eta_l$, global model $\mathbf{\Theta}_\mathcal{G}$;
}
\For{$t=0 \to T-1$}{
    Randomly sample a subset of clients $\mathbb{C}_M  \subseteq \mathbb{C}$\;
    \ForEach{client $\mathcal{C}_m \in \mathbb{C}_M$ \textbf{in parallel}}{
        Initialize $\mathbf{\Theta}_m^t$ via $\mathbf{\Theta}_\mathcal{G}^t$ from server\;
        Calculate $\mathcal{L}_\texttt{\textit{s}}$ and $\mathcal{L}_\texttt{\textit{u}}$ on $\mathbf{x}$ and $\mathbf{u}^\texttt{\textit{hc}}$ via Eq.~\ref{eq:kl_sup_loss_custom}\;
        Calculate $\mathcal{L}_\texttt{\textit{ICPL}}$ on $\mathbf{u}^\texttt{\textit{hc}}$ and $\mathbf{u}^\texttt{\textit{lc}}$ via Eq.~\ref{eq:icpl}\;
        $\mathcal{L}_\texttt{\textit{local}} \leftarrow \mathcal{L}_\texttt{\textit{s}} + \mathcal{L}_\texttt{\textit{u}} + \mathcal{L}_\texttt{\textit{ICPL}}$ 
        \;
        $\mathbf{\Theta}_m^{t+1} \leftarrow \mathbf{\Theta}_m^t - \eta_l \nabla_{\mathbf{\Theta}_m^t} \mathcal{L}_\texttt{\textit{local}}$\;
    }
    $\theta_\mathcal{G}^{t+1} {\leftarrow} \sum_{\mathcal{C}_m \in \mathbb{C}_M} \gamma_m \theta_m^{t+1},\mathbf{\Omega}_\mathcal{G} {\leftarrow} \sum_{\mathcal{C}_m \in \mathbb{C}_M} \gamma_m \omega_m^{t+1}$\;
    Server optimizes the global proxy $\mathbf{\Omega}_\mathcal{G}$ via Eq.~\ref{eq:loss_gpt},
    $\mathbf{\Omega}_\mathcal{G}^{t+1} \leftarrow \mathbf{\Omega}_\mathcal{G}\;$, thus $\mathbf{\Theta}_\mathcal{G}^{t+1} = \theta_\mathcal{G}^{t+1} \cup \mathbf{\Omega}_\mathcal{G}^{t+1}$;
}
\KwRet{$\mathbf{\Theta}_\mathcal{G}^T = \theta_\mathcal{G}^T \cup \mathbf{\Omega}_\mathcal{G}^T$}
\end{algorithm}
}

\begin{table*}[]
    \centering
    \caption{ Experimental results on CIFAR-10, CIFAR-100, SVHN and CINIC-10 under 20\% label. Bold text indicates the best result, and the last row presents the improvement of ProxyFL over the second best method.}
    \renewcommand{\arraystretch}{1}
    \resizebox{0.95\linewidth}{!}{
    \begin{tabular}{c|ccc|ccc|ccc|ccc}
    \hline\hline
         \multirow{2}{*}{\textbf{Methods}} & \multicolumn{3}{c}{CIFAR-10} & \multicolumn{3}{c}{CIFAR-100} & \multicolumn{3}{c}{SVHN} & \multicolumn{3}{c}{CINIC-10}\\
         & $\alpha=0.1$ & $\alpha=0.5$ & $\alpha=1$ & $\alpha=0.1$ & $\alpha=0.5$ & $\alpha=1$ & $\alpha=0.1$ & $\alpha=0.5$ & $\alpha=1$ & $\alpha=0.1$ & $\alpha=0.5$ & $\alpha=1$\\
    \hline\hline
        \multicolumn{1}{c}{\textbf{FL Methods}} & \multicolumn{12}{c}{} \\
    \hline\hline
        FedAvg \cite{mcmahan2017fedavg} & 86.37 & 87.06 & 87.97 & 45.72 & 46.57 & 47.55 & 88.37 & 89.05 & 89.97 & 66.24 & 68.29 & 69.21 \\
        FedProx \cite{li2020fedprox}  & 86.78 & 88.11 & 88.44 & 45.96 & 47.33 & 47.89 & 87.99 & 88.56 & 91.10 & 65.53 & 69.57 & 69.91 \\
        FedAvg-SL & 90.46 & 91.24 & 91.32 & 67.98 & 68.83 & 69.10 & 94.11 & 94.41 & 94.40 & 77.82 & 80.42 & 81.29 \\
        
    \hline\hline
        \multicolumn{1}{c}{\textbf{FL+SSL Methods}} & \multicolumn{12}{c}{} \\
    \hline\hline
        FixMatch-LPL & 87.22 & 89.61 & 89.23 & 56.80 & 57.35 & 57.59 & 93.66 & 94.11 & 94.21 & 72.51 & 75.14 & 76.03 \\
        FixMatch-GPL & 88.55 & 89.69 & 89.83 & 57.02 & 57.85 & 57.85 & 93.89 & 94.12 & 94.17 & 76.14 & 77.35 & 77.82 \\
        FedProx+FixMatch & 87.47 & 89.46 & 89.56 & 57.44 & 57.91 & 57.87 & 93.60 & 93.93 & 94.05 & 72.36 & 75.15 & 76.06 \\
        FedAvg+FlexMatch & 76.36 & 78.66 & 78.76 & 58.24 & 58.44 & 58.79 & 56.94 & 58.58 & 62.19 & 73.32 & 75.75 & 75.95 \\
    \hline\hline
        \multicolumn{1}{c}{\textbf{FSSL Methods}}& \multicolumn{12}{c}{} \\
    \hline\hline

        FedMatch \cite{jeong2020fedmatch} & 82.44 & 84.13 & 85.21 & 45.07 & 47.29 & 48.40 & 93.01 & 93.58 & 93.76 & 66.94 & 68.60 & 72.34 \\
        FedLabel \cite{cho2023fedlabel} & 87.37 & 88.86 & 88.93 & 58.63 & 58.98 & 59.23 & 93.44 & 94.38 & 94.59 & 60.13 & 67.30 & 72.22 \\
        FedLoke \cite{zhang2023fedloke} & 84.57 & 85.26 & 86.98 & 53.87 & 53.67 & 54.56 & 93.26 & 93.45 & 93.57 & 70.63 & 71.61 & 71.78 \\
        FedDure \cite{bai2024feddure} & 88.56 & 89.63 & 89.95 & 56.14 & 57.23 & 57.89 & 93.81 & 94.42 & 94.37 & 76.21 & 77.13 & 77.75 \\
        FedDB \cite{zhu2024feddb} & 85.19 & 86.36 & 86.65 & 52.81 & 54.62 & 55.48 & 93.22 & 93.50 & 94.27 & 74.18 & 75.00 & 75.65 \\
        SAGE \cite{liu2025sage} & 89.87 & 90.53 & 90.54 & 60.86 & 61.49 & 62.01 & 94.31 & 94.56 & 94.68 & 77.51 & 78.23 & 78.77 \\
        
        \multirow{2}{*}{\textbf{ProxyFL (ours)}} & \cellcolor[gray]{0.9}\textbf{90.97} & \cellcolor[gray]{0.9}\textbf{91.22} & \cellcolor[gray]{0.9}\textbf{91.51} & \cellcolor[gray]{0.9}\textbf{62.57} & \cellcolor[gray]{0.9}\textbf{63.09} & \cellcolor[gray]{0.9}\textbf{63.19} & \cellcolor[gray]{0.9}\textbf{95.03} & \cellcolor[gray]{0.9}\textbf{95.40} & \cellcolor[gray]{0.9}\textbf{95.34} & \cellcolor[gray]{0.9}\textbf{80.80} & \cellcolor[gray]{0.9}\textbf{81.02} & \cellcolor[gray]{0.9}\textbf{81.46} \\ 
        & \textcolor{RoyalBlue}{$\uparrow$ 1.10} & \textcolor{RoyalBlue}{$\uparrow$ 0.69} & \textcolor{RoyalBlue}{$\uparrow$ 0.97} & \textcolor{RoyalBlue}{$\uparrow$ 1.71} & \textcolor{RoyalBlue}{$\uparrow$ 1.60} & \textcolor{RoyalBlue}{$\uparrow$ 1.18} & \textcolor{RoyalBlue}{$\uparrow$ 0.72} & \textcolor{RoyalBlue}{$\uparrow$ 0.84} & \textcolor{RoyalBlue}{$\uparrow$ 0.66} & \textcolor{RoyalBlue}{$\uparrow$ 3.29} & \textcolor{RoyalBlue}{$\uparrow$ 2.79} & \textcolor{RoyalBlue}{$\uparrow$ 2.69} \\
    \hline\hline
    \end{tabular}
    }
    \label{tab:comparison_20}
\end{table*}

\section{Theoretical Proofs}
\label{appendix:proof}
In this section, we provide the convergence analysis for the bi-level optimizations of ProxyFL: Global Proxy Tuning (GPT) and Indecisive-Categories Proxy Learning (ICPL). Our proofs are based on the standard assumptions in the non-convex optimization.
\subsection{Convergence of Global Proxy Tuning}
\label{sec:gpt_proof}
Our GPT module is a global optimization process executed on the server. In each communication round, the server collects the local proxies $\{\omega_m\}_{m=1}^{M}$ from clients and then optimizes the global proxies $\mathbf{\Omega}_\mathcal{G}$ by minimizing the loss function $\mathcal{L}_\texttt{\textit{GPT}}$. First, we give: 
\begin{theorem}
\label{theory:GPT}
\textup{\textbf{(Convergence of Global Proxy Tuning)}} Assume that the loss function $\mathcal{L}_\texttt{\textit{GPT}}$ is L-Lipschitz and bounded below, where $\mathcal{L}_\texttt{\textit{GPT}}$ is related to $\mathbf{\Omega}_\mathcal{G}$. By optimizing the global proxies $\mathbf{\Omega}_\mathcal{G}$ via gradient descent with learning rate $\eta_\mathcal{G}$ such that $0<\eta_\mathcal{G} \le \frac{1}{L_\mathcal{G}}$, the optimization process converges to a stationary point. I.e.,
\begin{align}
    \lim _{Q \rightarrow \infty} \frac{1}{Q} \sum_{q=0}^{Q-1} \mathbb{E}\left[\left\|\nabla_{\mathbf{\Omega}_{\mathcal{G}}} \mathcal{L}_\texttt{\textit{GPT}}\left(\mathbf{\Omega}_{\mathcal{G}}^{q}\right)\right\|^{2}\right]=0,
\end{align}
where $Q$ is the number of proxy tuning steps on the server. 
\end{theorem}
Then we provide a specific proof for Theorem~\ref{theory:GPT}. According to the descent lemma for L-smooth functions, we have:
\begin{align}
\label{eq:GPT_d_lemma}
    \mathcal{L}_\texttt{\textit{GPT}}(\mathbf{\Omega}_\mathcal{G}^{q+1})\leq & \mathcal{L}_\texttt{\textit{GPT}}(\mathbf{\Omega}_\mathcal{G}^q) +\langle\nabla_{\mathbf{\Omega}_\mathcal{G}}\mathcal{L}_\texttt{\textit{GPT}}(\mathbf{\Omega}_\mathcal{G}^q),\mathbf{\Omega}_\mathcal{G}^{q+1}-\mathbf{\Omega}_\mathcal{G}^q\rangle \nonumber \\ &+\frac{L_\mathcal{G}}{2}||\mathbf{\Omega}_\mathcal{G}^{q+1}-\mathbf{\Omega}_\mathcal{G}^q||^2.
\end{align}
According to the gradient-descent formula $\mathbf{\Omega}_\mathcal{G}^{q+1}=\mathbf{\Omega}_\mathcal{G}^q-\eta_\mathcal{G}\nabla_{\mathbf{\Omega}_\mathcal{G}}\mathcal{L}_\texttt{\textit{GPT}}(\mathbf{\Omega}_\mathcal{G}^q)$, Eq.~\ref{eq:GPT_d_lemma} can be re-written as:

\begin{align}
\label{eq:gpt_grad_des}
    \mathcal{L}_\texttt{\textit{GPT}}(\mathbf{\Omega}_\mathcal{G}^{q+1}) \leq & \mathcal{L}_\texttt{\textit{GPT}}(\mathbf{\Omega}_\mathcal{G}^{q})-\eta_\mathcal{G}||\nabla_{\mathbf{\Omega}_\mathcal{G}}\mathcal{L}_\texttt{\textit{GPT}}(\mathbf{\Omega}_\mathcal{G}^{q})||^{2} \\ \nonumber & +\frac{L_\mathcal{G}\eta_\mathcal{G}^{2}}{2}||\nabla_{\mathbf{\Omega}_\mathcal{G}}\mathcal{L}_\texttt{\textit{GPT}}(\mathbf{\Omega}_\mathcal{G}^{q})||^{2}
\end{align}
Then, the right side will be: 
\begin{align}
\label{eq:gpt_grad_des_2}
    \mathcal{L}_\texttt{\textit{GPT}}(\mathbf{\Omega}_\mathcal{G}^{q+1}) \leq \mathcal{L}_\texttt{\textit{GPT}}(\mathbf{\Omega}_\mathcal{G}^{q})-\eta_\mathcal{G}(1-\frac{L_\mathcal{G}\eta_\mathcal{G}}{2})||\nabla_{\mathbf{\Omega}_\mathcal{G}}\mathcal{L}_\texttt{\textit{GPT}}(\mathbf{\Omega}_\mathcal{G}^{q})||^{2} 
\end{align}
Let the learning rate $\eta_\mathcal{G} \le \frac{1}{L_\mathcal{G}}$, such that $1-\frac{L_\mathcal{G}\eta_\mathcal{G}}{2} \ge \frac{1}{2}$. Thus, Eq.~\ref{eq:gpt_grad_des} can be simplified to:
\begin{align}
\label{eq:gpt_1/2}
    \mathcal{L}_\texttt{\textit{GPT}}(\mathbf{\Omega}_\mathcal{G}^{q+1})\leq\mathcal{L}_\texttt{\textit{GPT}}(\mathbf{\Omega}_\mathcal{G}^q)-\frac{\eta_\mathcal{G}}{2}||\nabla_{\mathbf{\Omega}_\mathcal{G}}\mathcal{L}_\texttt{\textit{GPT}}(\mathbf{\Omega}_\mathcal{G}^q)||^2
\end{align}
Rearranging the terms, we get:
\begin{align}
\label{eq:gpt_re-arrange}
    ||\nabla_{\mathbf{\Omega}_\mathcal{G}}\mathcal{L}_\texttt{\textit{GPT}}(\mathbf{\Omega}_\mathcal{G}^q)||^2\leq\frac{2}{\eta_\mathcal{G}}(\mathcal{L}_\texttt{\textit{GPT}}(\mathbf{\Omega}_\mathcal{G}^q)-\mathcal{L}_\texttt{\textit{GPT}}(\mathbf{\Omega}_\mathcal{G}^{q+1}))
\end{align}
Summing both the left and right sides of Eq.~\ref{eq:gpt_re-arrange} from $q=0$ to $Q-1$, we have:
\begin{align}
    \sum_{q=0}^{Q-1}\|\nabla_{\mathbf{\Omega}_\mathcal{G}}\mathcal{L}_\texttt{\textit{GPT}}(\mathbf{\Omega}_\mathcal{G}^q)\|^2 & \leq \underbrace{\frac{2}{\eta_\mathcal{G}}\sum_{q=0}^{Q-1}(\mathcal{L}_\texttt{\textit{GPT}}(\mathbf{\Omega}_\mathcal{G}^q)-\mathcal{L}_\texttt{\textit{GPT}}(\mathbf{\Omega}_\mathcal{G}^{q+1}))} \nonumber \\
    & = \frac{2}{\eta_\mathcal{G}}(\mathcal{L}_\texttt{\textit{GPT}}(\mathbf{\Omega}_\mathcal{G}^0)-\mathcal{L}_\texttt{\textit{GPT}}(\mathbf{\Omega}_\mathcal{G}^Q))
\end{align}
Since we adopt InfoNCE loss \cite{oord2018infonce} for $\mathcal{L}_\texttt{\textit{GPT}}$ with its lowerbound $0$, we thus have:
\begin{align}
    \sum_{q=0}^{Q-1}\|\nabla_{\mathbf{\Omega}_\mathcal{G}}\mathcal{L}_\texttt{\textit{GPT}}(\mathbf{\Omega}_\mathcal{G}^q)\|^2 \leq \frac{2}{\eta_\mathcal{G}}\mathcal{L}_\texttt{\textit{GPT}}(\mathbf{\Omega}_\mathcal{G}^0)
\end{align}
Dividing both sides by $Q$ and taking the limit, we have:
\begin{align}
\label{eq:gpt_final}
    \lim_{Q\to\infty}\frac{1}{Q}\sum_{q=0}^{Q-1}||\nabla_{\mathbf{\Omega}_\mathcal{G}}\mathcal{L}_\texttt{\textit{GPT}}(\mathbf{\Omega}_\mathcal{G}^q)||^2\leq\lim_{Q\to\infty}\frac{2\mathcal{L}_\texttt{\textit{GPT}}(\mathbf{\Omega}_\mathcal{G}^0)}{\eta_\mathcal{G}Q}=0
\end{align}
Since the squared norm of the gradient (\emph{i.e., the left-hand side of Eq.~\ref{eq:gpt_final}}) is non-negative, \textbf{hence we have proven Theorem~\ref{theory:GPT}.}

\subsection{Convergence of Local Training with ICPL}
The ICPL module is executed during the local training on each client. The total local loss is denoted as $\mathcal{L}_\texttt{\textit{local}}=\mathcal{L}_\texttt{\textit{s}}+\alpha\mathcal{L}_\texttt{\textit{u}}+\beta\mathcal{L}_\texttt{\textit{ICPL}}$, with optimization parameters $\mathbf{\Theta}_k=\theta_k \cup \omega_k.$
\begin{theorem}
\label{theory:ICPL}
\textup{\textbf{(Convergence of Local Training with ICPL)}} Suppose that the total local loss function $\mathcal{L}_k$ for client k is L-smooth and bounded below, where $\mathcal{L}_k$ is related to $\mathbf{\Theta}_k$. By optimizing the local model parameters $\mathbf{\Theta}_k$ via gradient descent with learning rate $\eta_l$ such that $0 < \eta_l \le \frac{1}{L_k}$ , the optimization
process converges to a stationary point. I.e.,
\begin{align}
    \lim_{E\to\infty}\frac{1}{E}\sum_{e=0}^{E-1}\mathbb{E}[||\nabla_{\mathbf{\Theta}_k}\mathcal{L}_k(\mathbf{\Theta}_k^e)||^2]=0
\end{align}
where $E$ is the number of local training epochs.

\end{theorem}
Similar to Sec.~\ref{sec:gpt_proof}, following the descent lemma and the gradient-descent update formula, we can similarly derive:
\begin{align}
    \mathcal{L}_k(\mathbf{\Theta}_k^{e+1})\leq\mathcal{L}_k(\mathbf{\Theta}_k^e)-\eta_l(1-\frac{L_k\eta_l}{2})\|\nabla_{\mathbf{\Theta}_k}\mathcal{L}_k(\mathbf{\Theta}_k^e)\|^2
\end{align}
By setting a local learning rate $\eta_l \le \frac{1}{L_k}$, we obtain:
\begin{align}
\label{eq:icpl_re}
    \sum_{e=0}^{E-1}||\nabla_{\Theta_k}\mathcal{L}_k(\mathbf{\Theta}_k^e)||^2&\leq\frac{2}{\eta_l}(\mathcal{L}_k(\mathbf{\Theta}_k^0)-\mathcal{L}_k(\mathbf{\Theta}_k^E)) \nonumber \\ &\leq \frac{2}{\eta_l}\mathcal{L}_k(\mathbf{\Theta}_k^0)
\end{align}
Taking the limit to Eq.~\ref{eq:icpl_re} and knowing that the squared norm is non-negative, we have:
\begin{align}
    0 \leq \lim_{E\to\infty}\frac{1}{E}\sum_{e=0}^{E-1}||\nabla_{\mathbf{\mathbf{\Theta}}_k}\mathcal{L}_k(\mathbf{\Theta}_k^e)||^2 \leq \lim_{E\to\infty}\frac{2}{\eta_lE}\mathcal{L}_k(\mathbf{\Theta}_k^0) = 0
\end{align}
Thus,
\begin{align}
     \lim_{E\to\infty}\frac{1}{E}\sum_{e=0}^{E-1}||\nabla_{\mathbf{\Theta}_k}\mathcal{L}_k(\mathbf{\Theta}_k^e)||^2 = 0
\end{align}
\textbf{So we have proven Theorem~\ref{theory:ICPL}.}

\section{Additional Empirical Analysis}
\label{appendix:add_exp}
We provide some additional experiments to complement Sec.~\ref{sec:experiments} of our main paper, thereby offering a more robust demonstration of our proposed ProxyFL.

\begin{table}[t]
    \centering
    \caption{The communication costs per round between different methods in our experimental setting as SAGE \cite{liu2025sage}. $\mathbf{\Theta}$ represents model parameters. $\sigma$ and $\psi$ denotes the parameters after sparsely decomposition in FedMatch \cite{jeong2020fedmatch}, and $H$ means the number of helper agents. $C$ and $d$ means respectively the number of categories and feature dimension.}
    \resizebox{0.95\linewidth}{!}{
    \begin{tabular}{ccc}
    \hline\hline
         &  Formulation & Practice \\
    \hline\hline
    FixMatch+FSSL & $\sum_{m=1}^{M}|\mathbf{\Theta}| + |\mathbf{\Theta}|$ & 22.08 M\\
    \multirow{2}{*}{FedMatch} & $\sum_{i=1}^M(|\Delta\sigma_i|+|\Delta\psi_i|+C)+$ &  \multirow{2}{*}{33.12 M}\\
    & $(|\Delta\sigma_\mathcal{G}|+|\Delta\psi_\mathcal{G}|+H\times|\psi_h|))$ & \\
    FedLabel & $\sum_{m=1}^{M}|\mathbf{\Theta}| + |\mathbf{\Theta}|$ & 22.08 M\\
    FedDure & $\sum_{m=1}^{M}|\mathbf{\Theta}| + |\mathbf{\Theta}|$ & 22.08 M\\
    FedDB & $\sum_{i=1}^M(|\mathbf{\Theta}|+C)+|\mathbf{\Theta}|$ & 22.08 M\\
    SAGE & $\sum_{m=1}^{M}|\mathbf{\Theta}| + |\mathbf{\Theta}|$ & 22.08 M\\
    Proto. + FSSL & $\sum_{m=1}^{M}|\mathbf{\Theta}| + |\mathbf{\Theta}| + C\times d$ & 22.90 M\\
    ProxyFL & \cellcolor[gray]{0.9}{$\sum_{i=1}^M(|\mathbf{\Theta}|+C)+(|\mathbf{\Theta}|+C)$} & \cellcolor[gray]{0.9}{22.08 M}\\
    \hline\hline
    \end{tabular}
    }
    \label{tab:comm_cost}
\end{table}

\begin{table}[t]
    \centering
    \caption{Comparison of convergence rates between ProxyFL and other baseline methods on CIFAR100 dataset with $\alpha = 0.5$ (the upper part) and $\alpha = 1$ (the lower part).}
    \renewcommand{\arraystretch}{1}
    \resizebox{1\linewidth}{!}{
    \begin{tabular}{ccccccc}
    \hline\hline
        Acc. & \multicolumn{2}{c}{30\%} & \multicolumn{2}{c}{40\%} & \multicolumn{2}{c}{50\%} \\
        Method & Round$\downarrow$ & Speedup$\uparrow$ & Round$\downarrow$ & Speedup$\uparrow$ & Round$\downarrow$ & Speedup$\uparrow$ \\
    \hline\hline
        LPL & 121 & $\times$1.00 & 221 & $\times$1.00 & 546 & $\times$1.00 \\
        GPL & 113 & $\times$1.07 & 210 & $\times$1.05 & 419 & $\times$1.30 \\
        FedLabel & 83 & $\times$1.46 & 160 & $\times$1.38 & 366 & $\times$1.49 \\
        FedDB & 94 & $\times$1.29 & 205 & $\times$1.08 & 492 & $\times$1.11 \\
        FedDure & 110 & $\times$1.10 & 222 & $\times$1.00 & 552 & $\times$0.99 \\
        SAGE & 55 & $\times$2.20 & 105 & $\times$2.10 & 241 & $\times$2.27 \\
        \textbf{ProxyFL} & \cellcolor[gray]{0.9}\textbf{44} & $\times$\cellcolor[gray]{0.9}\textbf{2.75} & \cellcolor[gray]{0.9}\textbf{79} & $\times$\cellcolor[gray]{0.9}\textbf{2.80} & \cellcolor[gray]{0.9}\textbf{155} & $\times$\cellcolor[gray]{0.9}\textbf{3.52} \\
    \hline\hline
        LPL & 118 & $\times$1.00 & 267 & $\times$1.00 & 527 & $\times$1.00 \\
        GPL & 94 & $\times$1.26 & 183 & $\times$1.46 & 390 & $\times$1.35 \\
        FedLabel & 91 & $\times$1.30 & 164 & $\times$1.63 & 341 & $\times$1.55 \\
        FedDB & 103 & $\times$1.15 & 237 & $\times$1.13 & 418 & $\times$1.26 \\
        FedDure & 95 & $\times$1.24 & 182 & $\times$1.47 & 450 & $\times$1.17 \\
        SAGE & 56 & $\times$2.11 & 112 & $\times$2.38 & 242 & $\times$2.18 \\
        \textbf{ProxyFL} & \cellcolor[gray]{0.9}\textbf{45} & $\times$\cellcolor[gray]{0.9}\textbf{2.62} & \cellcolor[gray]{0.9}\textbf{83} & $\times$\cellcolor[gray]{0.9}\textbf{3.22} & \cellcolor[gray]{0.9}\textbf{157} & $\times$\cellcolor[gray]{0.9}\textbf{3.36} \\
    \hline\hline
    \end{tabular}
    }
    \label{tab:convergence_rate_0.5_1}
\end{table}

\subsection{Labeling Ratio} 
To validate the effectiveness of our ProxyFL, we give a comparison between ProxyFL and other state-of-the-art methods with a 10\% labeling ratio in Tab.~\ref{tab:comparison} of main paper. Here, we further study the robustness of ProxyFL by comparing ProxyFL with other methods at 20\% labeling ratio. As shown in Tab.~\ref{tab:comparison_20}, ProxyFL consistently achieves better performance across different labeling ratios than others.

\subsection{Communication Costs}
We calculate the communication cost in both theory and practice. As shown in Tab.~\ref{tab:comm_cost}, FedMatch \cite{jeong2020fedmatch} needs the most communication costs of some auxiliary parameters since they additionally uploads the model embedding for KD-Tree reconstruction and downloads $H$ helper agents to facilitate local training. Incorporating `prototypes' into FSSL will bring extra costs about $0.82$ M as $\sum_{m=1}^{M} C \times d$ and high-dimensional features (prototypes) poses the risk of reversely reconstruction \cite{dosovitskiy2016inverting,melis2019exploiting}. As shown in Tab.~\ref{tab:comm_cost}, compared to \cite{cho2023fedlabel,bai2024feddure,liu2025sage}, the costs of $\mathcal{P}'_\mathcal{G}(\mathbf{Y})$ in ProxyFL could be negligible ($\sum_{m=1}^{M}C+C$, approximately $0.0016$ M, similar to \cite{zhu2024feddb}) and our method achieves much higher accuracy than them. In summary, our ProxyFL achieves higher accuracy while preserving a reasonable communication efficiencies and our prior design has lower privacy-leakage risk \cite{zhu2024feddb} than raw samples or features.

\subsection{Convergence Rate}
In the 2) of Sec.~\ref{sec:empirical} of the main paper, we conduct experiments under the $\alpha = 0.1$ setting, where the ProxyFL method significantly improve model convergence speed and test accuracy. Here, we provide a detailed comparison of SAGE and other methods under varying levels of heterogeneity. As demonstrated in Tab.~\ref{tab:convergence_rate_0.5_1} and theoretical proofs of Sec.~\ref{appendix:proof}, ProxyFL could achieve substantial acceleration in early stages and ensure final convergence under different levels of heterogeneity.

\subsection{ProxyFL as a Plug-in Approach}
The GPT and ICPL components in our ProxyFL can be considered as a optimization mechanism of category distribution for FSSL, allowing integration into other FSSL approaches. As present in Tab.~\ref{tab:plugin}, ProxyFL consistently improves the performance of existing FSSL methods. With a proxy-guided mechanism, our approach alleviates both internal and external heterogeneity locally and globally.
\begin{table}[]
    \centering
    \caption{Performance gains brought by ProxyFL as a plugin to other baseline methods.}
    \renewcommand{\arraystretch}{1}
    \resizebox{0.95\linewidth}{!}{
    \begin{tabular}{ccccc}
    \hline\hline
        Methods          & CIFAR-10 & CIFAR-100 & SVHN  & CINIC-10 \\
    \hline\hline
        FixMatch    &  82.98 & 49.32    &  89.68  & 68.02  \\ 
        \multirow{2}{*}{+ProxyFL} & \cellcolor[gray]{0.9}{88.56} &  \cellcolor[gray]{0.9}{57.50}  & \cellcolor[gray]{0.9}{95.09} & \cellcolor[gray]{0.9}{77.98} \\ 
        &   \textcolor{RoyalBlue}{$\uparrow$ 5.58}  & \textcolor{RoyalBlue}{$\uparrow$ 8.18}  &  \textcolor{RoyalBlue}{$\uparrow$ 5.41} & \textcolor{RoyalBlue}{$\uparrow$ 9.96} \\ 
    \hline\hline
        FedLabel & 62.85 & 50.88 & 89.31 & 67.64\\
                \multirow{2}{*}{+ProxyFL} &    \cellcolor[gray]{0.9}{88.55} &  \cellcolor[gray]{0.9}{56.88}  & \cellcolor[gray]{0.9}{94.94} &  \cellcolor[gray]{0.9}{76.35}\\ 
        &   \textcolor{RoyalBlue}{$\uparrow$ 25.70} &   \textcolor{RoyalBlue}{$\uparrow$ 6.00} & \textcolor{RoyalBlue}{$\uparrow$ 5.63} &  \textcolor{RoyalBlue}{$\uparrow$ 8.71}\\ 
    \hline\hline
        SAGE & 87.05 & 54.18 & 93.85 & 74.59 \\
                        \multirow{2}{*}{+ProxyFL} &  \cellcolor[gray]{0.9}{88.29} & \cellcolor[gray]{0.9}{57.10} & \cellcolor[gray]{0.9}{95.01} &  \cellcolor[gray]{0.9}{77.58} \\ 
        &    \textcolor{RoyalBlue}{$\uparrow$ 1.24} &  \textcolor{RoyalBlue}{$\uparrow$ 2.92} & \textcolor{RoyalBlue}{$\uparrow$ 1.16} &    \textcolor{RoyalBlue}{$\uparrow$ 2.99}\\ 
    \hline\hline
    \end{tabular}
    }
    \label{tab:plugin}
\end{table}
\begin{figure}
    \centering
    \includegraphics[width=1\linewidth]{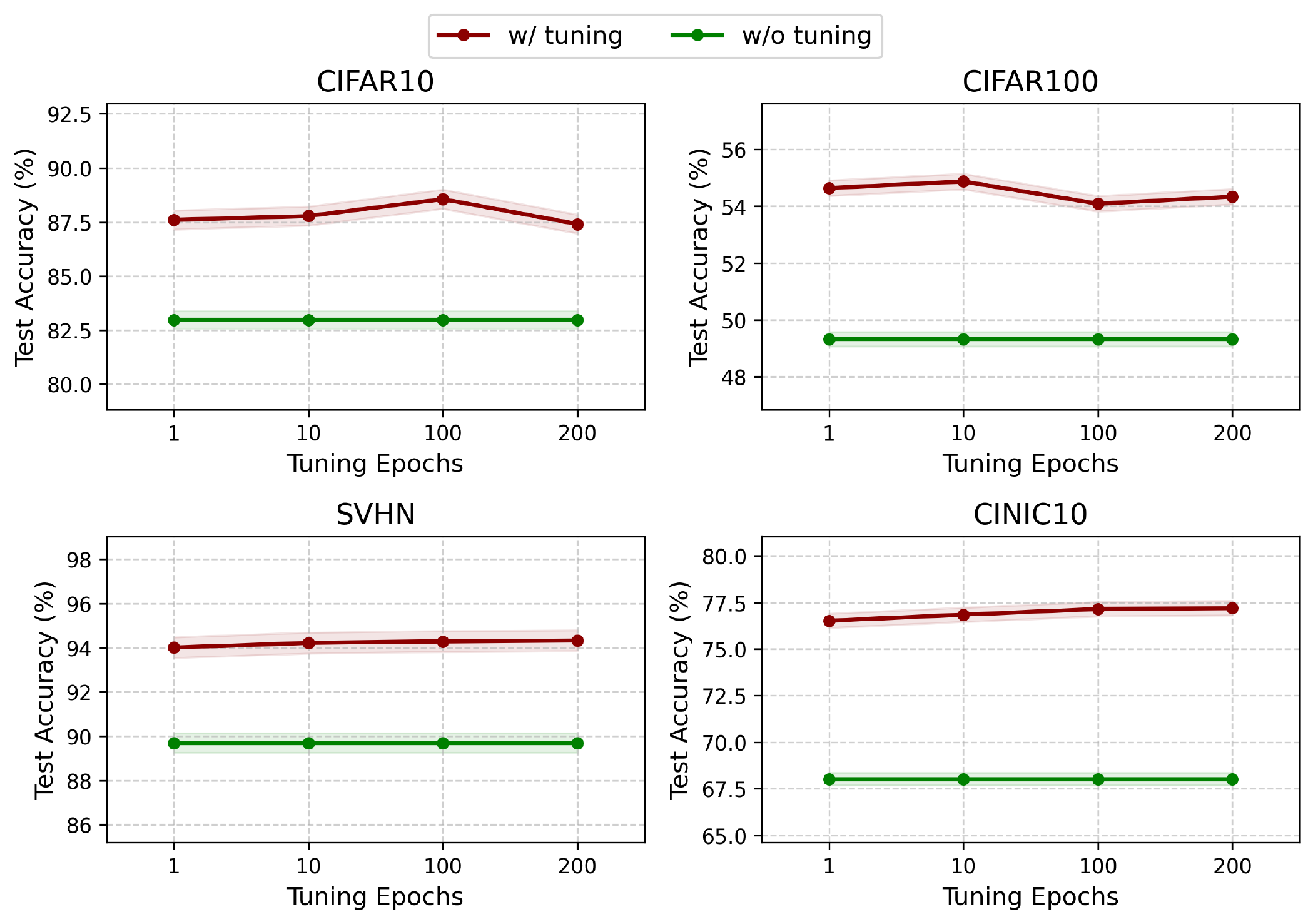}
    \caption{Ablation on the numbers of epoch for Global Proxy Tuning.}
    \label{fig:server_epochs}
\end{figure}
\subsection{Ablation Study on Tuning Epochs}
We conduct experiments for server-tuning epochs, \emph{i.e.}, the hyper-parameter $Q$. As shown in Fig.~\ref{fig:server_epochs}, our GPT module achieves best performance when $Q\approx100$ across most datasets. But, for CIFAR100 dataset, it requires fewer tuning epochs (\emph{i.e.}, $Q \approx 10$) and exhibits overfitting when $Q$ increases. We analyze that, since CIFAR100 dataset has more categories ($\times10$ times category-number than other datasets) and thus has more tuning samples, it will converge much faster than other datasets. 
Thus, corresponding to Sec.~\ref{sec:setup} of our main paper, we claim that the number of tuning epochs is set to 10 for CIFAR-100 dataset and 100 for the other datasets.

\subsection{Experimental Setting and other Details}
As claimed in Sec.~\ref{sec:setup} of our main paper, \textbf{we strictly follow the experimental setup of SAGE \cite{liu2025sage}.} We conduct our experiments on the same datasets with consistent client-splitting strategies as SAGE. For the hyper-parameter `the communication rounds $T$', we follow SAGE to set $300,500,150,400$ rounds for CIFAR10, CIFAR100, SVHN and CINIC10 dataset, respectively. The local learning rate is set to $0.1$ and the confidence threshold $\tau$ for pseudo-labeling set to 0.95, \emph{etc.} In summary, unless otherwise specified, our experimental details are consistent with SAGE.

\end{document}